\pdfoutput=1

\documentclass[11pt]{article}

\usepackage{style/acl}

\usepackage{times}
\usepackage{latexsym}

\usepackage[T1]{fontenc}

\usepackage[utf8]{inputenc}

\usepackage{microtype}

\usepackage{amsmath}
\usepackage{multirow}
\usepackage{multicol}
\usepackage{xcolor}
\usepackage{graphicx}
\usepackage{booktabs}
\usepackage{longtable}
\usepackage{microtype}
\usepackage[ruled,linesnumbered,vlined]{algorithm2e}
\usepackage{adjustbox}
\usepackage{rotating}
\usepackage{style/slashbox}
\usepackage{soul}
\usepackage[normalem]{ulem}
\usepackage{paralist}
\usepackage{float}

\SetKwInput{KwInput}{Input}                
\SetKwInput{KwOutput}{Output}              
\DeclareMathOperator*{\argmax}{argmax}

\usepackage{lipsum}

\definecolor{ForestGreen}{RGB}{34,139,34}

\title{CrossSum: Beyond English-Centric Cross-Lingual Summarization for 1,500+ Language Pairs}

\author{
Abhik Bhattacharjee$^1$\thanks{~ These authors contributed equally to this work.} , Tahmid Hasan$^1$\footnotemark[1] , Wasi Uddin Ahmad$^2$,\\
\textbf{Yuan-Fang Li}$^3$, \textbf{Yong-Bin Kang}$^4$, \textbf{Rifat Shahriyar}$^1$\\ [3pt]
Bangladesh University of Engineering and Technology (BUET)$^1$, University of California,\\
Los Angeles$^2$, Monash University$^3$, Swinburne University of Technology$^4$\\ [3pt]
\texttt{\{tahmidhasan,rifat\}@cse.buet.ac.bd}, \texttt{abhik@ra.cse.buet.ac.bd}\\}

\date{}

\begin{document}

\maketitle

\setlength{\abovedisplayskip}{5pt}
\setlength{\belowdisplayskip}{5pt}

\begin{abstract}
We present CrossSum, a large-scale cross-lingual summarization dataset comprising 1.68 million article-summary samples in 1,500+ language pairs. We create CrossSum by aligning parallel articles written in different languages via cross-lingual retrieval from a multilingual abstractive summarization dataset and perform a controlled human evaluation to validate its quality. We propose a multistage data sampling algorithm to effectively train a cross-lingual summarization model capable of summarizing an article in any target language. We also introduce LaSE, an embedding-based metric for automatically evaluating model-generated summaries. LaSE is strongly correlated with ROUGE and, unlike ROUGE, can be reliably measured even in the absence of references in the target language. Performance on ROUGE and LaSE indicate that our proposed model consistently outperforms baseline models. To the best of our knowledge, CrossSum is the largest cross-lingual summarization dataset and the first ever that is not centered around English. We are releasing the dataset, training and evaluation scripts, and models to spur future research on cross-lingual summarization. The resources can be found at \protect\url{https://github.com/csebuetnlp/CrossSum}.
\end{abstract}

\begin{figure}[ht]
 \centering
 \includegraphics[width=0.49\textwidth]{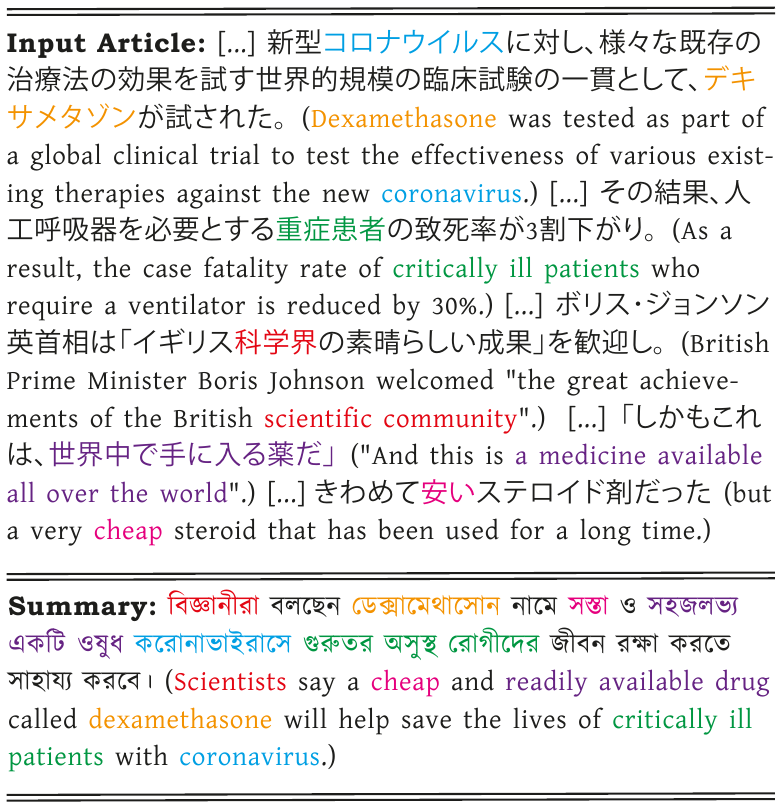}
 \caption{A sample article-summary pair from CrossSum, the article is written in Japanese, and the summary is in Bengali. We translate the texts to English inside parentheses for better understanding. Words and phrases of the article relevant to the summary are color-coded.}
 \label{fig:example}
\end{figure}

\section{Introduction}

Cross-lingual summarization (hereinafter XLS) is the task of generating a summary in a target language given a source text in another language. The task is challenging as it combines summarization and translation in one task, both challenging tasks in their own right. Earlier approaches to XLS thus employed pipeline methods such as translate-then-summarize \citep{leuski2003cross} and summarize-then-translate \citep{wan-etal-2010-cross}. Not only are they computationally expensive, having to use multiple models, but these approaches also suffer from error-propagation \citep{zhu-etal-2019-ncls} from one model to another, degrading the overall performance. 

The success of sequence-to-sequence (seq2seq) models \citep{cho2014learning, sutskever2014sequence} and the advances in Transformer-based models \citep{vaswani2017attention} have aided in the emergence of end-to-end methods that can perform XLS with one single model \citep{zhu-etal-2019-ncls, cao2020multisumm}. The availability of XLS datasets \citep{ladhak-etal-2020-wikilingua, perez-beltrachini-lapata-2021-models} has also helped this task gain popularity in recent times. However, they cover only a few languages, contain a small number of samples for training and evaluation, or use English as the pivot language (i.e., the target language always remains English), thereby limiting their applicability to a great extent.

To democratize XLS beyond high-resource languages, in this work, we introduce \textbf{CrossSum}, a large-scale XLS dataset containing 1.68 million article-summary samples in 1,500+ language pairs. We align parallel articles\footnote{We re-purpose the terminology of parallel corpus here.} written in different languages via cross-lingual retrieval from the multilingual XL-Sum \citep{hasan-etal-2021-xl} dataset. We introduce and rigorously study the notions `\textit{induced pairs}' and `\textit{implicit leakage}' to increase the coverage of the dataset while at the same time ensuring maximum quality. We also perform a controlled human evaluation of CrossSum spanning nine languages from high- to low-resource and show that the alignments are highly accurate. 

We design \textbf{MLS}, a multistage language sampling algorithm, for successfully training models that can generate a summary in any target language for an input article in any source language, both from a set of languages present in the training dataset. For the first time, we perform XLS with CrossSum on a broad and diverse set of languages without relying on English as the standalone pivot, consistently outperforming many-to-one and one-to-many models, as well as summarize-then-translate baselines. 

We propose \textbf{LaSE}, an embedding-based metric for evaluating summaries when reference summaries may not be available in the target language but may be available in another language, potentially opening new doors for evaluating low-resource languages. Furthermore, we demonstrate the reliability of LaSE by its high correlation with ROUGE \cite{lin2004rouge}, the de-facto metric for evaluating text summarization systems.

To the best of our knowledge, CrossSum is the largest publicly available abdtractive XLS dataset, both in terms of the number of samples and the number of language pairs. We are releasing the dataset, training and evaluation scripts, and models hoping that these resources will encourage the community to push the boundaries of XLS beyond English and other high-resource languages.

\section{The CrossSum Dataset}\label{sec:cross}

The most straightforward way of curating a high-quality XLS dataset is via crowd-sourcing \citep{nguyen-daume-iii-2019-global}. However, it may be difficult to find crowd workers having professional command over low-resource languages or distant language pairs. Moreover, scalability issues might arise due to the time and budget constraints for crowd-sourcing. Therefore, synthetic \citep{zhu-etal-2019-ncls} and automatic methods \citep{ladhak-etal-2020-wikilingua, perez-beltrachini-lapata-2021-models} have gained traction over crowd-sourcing.

Automatic curation of an XLS dataset is simply to pair an article \texttt{A} in a source language with the summary of a parallel article \texttt{B} written in a different target language (Figure \ref{fig:example}), assuming the availability of a multilingual dataset having identical contents in different languages. Two contemporary works have compiled large-scale multilingual summarization datasets, namely XL-Sum \citep{hasan-etal-2021-xl} (1.35M samples in 45 languages)  and MassiveSumm \citep{varab-schluter-2021-massivesumm} (28.8M samples in 92 languages). Though substantially larger than the other, MassiveSumm is not publicly available. Since public availability is crucial for promoting open research, we opted for XL-Sum, distributed under a non-commercial license. Additionally, all articles of XL-Sum are crawled from a single source, BBC News. We observed that BBC publishes similar news content in different languages and follow similar summarization strategies. Hence adopting XL-Sum would increase the quality and quantity of the article-summary pairs.

Unlike previous automatic methods, there are no explicit links between parallel articles in XL-Sum. Fortunately, language-agnostic sentence representations \citep{artetxe-schwenk-2019-margin, feng-etal-2022-language} have achieved state-of-the-art results in cross-lingual text mining \citep{artetxe2019massively}, and hence, we use them to search identical contents across languages. For simplicity\footnote{The entire procedure is described in Appendix \ref{sec:LaBSE}.}, we perform the search over summaries only. To ensure maximum quality, we set two conditions for a summary $S_A$ in language \texttt{A} to be aligned with another summary $S_B$ in language \texttt{B}:
\begin{compactenum}
    \item $S_B$ must be the nearest neighbor of $S_A$ among all summaries in \texttt{B}, and vice-versa.
    \item The similarity between $S_A$ and $S_B$ must be above the threshold, $\tau$.
\end{compactenum}

The similarity of a summary pair is measured by the inner product of their Language-agnostic BERT Sentence Embeddings (LaBSE) \citep{feng-etal-2022-language} (a unit vector for an input text sequence). We empirically set the similarity threshold as the average over all languages that maximized their respective $F_1$ score ($\tau = 0.7437$) in the BUCC mining tasks \citep{zweigenbaum2017overview}.\footnote{Around 90\% $F_1$ is achieved using LaBSE in BUCC, hence not all CrossSum alignments will be correct. Therefore, in the following section, we further assess the quality of the alignments using human evaluation.} 

\begin{figure*}[!tbh]
\centering
\includegraphics[width=\textwidth]{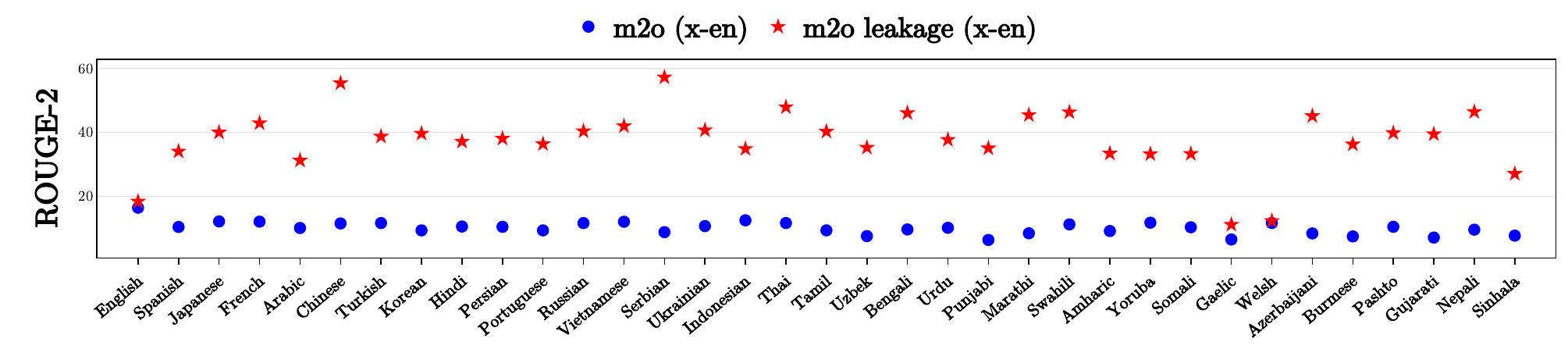}
\caption{Training on the dataset respecting the original XL-Sum splits causes unusually high ROUGE scores (marked red) in many-to-one models due to implicit data leakage. Therefore, we redid the splits taking the issue into account, and consequently, models trained on the new set (marked blue) do not exhibit any unusual spike.}
\label{fig:leak}
\end{figure*}

\paragraph{Induced Pairs} We observed that many summary pairs, despite being nearest neighbors in their language pairs, were filtered out because of the threshold $\tau$. Although interestingly, both were aligned with the same summary in a different language. Moreover, these pairs are prevalent if their languages are distant or low-resource. LaBSE uses contrastive learning \citep{guo-etal-2018-effective, ijcai2019-746} to rank parallel sentences over non-parallels. Since parallel pairs are mostly found for high-resource and linguistically close languages, we hypothesize that LaBSE fails to assign high similarity to sentences from languages that are not.

To include these pairs into CrossSum, we introduce the notion `\textit{induced pairs}.' Formally, two summaries $S_A, S_B$ in languages \texttt{A}, \texttt{B} are induced pairs if they are nearest neighbors of each other in \texttt{A}, \texttt{B}, their similarity score is below $\tau$, and both are aligned with $S_C$ in language \texttt{C}, or through a chain of aligned pairs $(S_A, S_C), (S_C, S_D), \cdots, (S_Y, S_Z), (S_Z, S_B)$ in languages $\{$\texttt{C}, \texttt{D}, $\cdots$, \texttt{Y}, \texttt{Z}$\}$.

We thus incorporate the induced pairs into CrossSum through a simple graph-based algorithm. First, we represent all summaries as vertices in a graph and draw an edge between two vertices if the summaries are aligned. Then we find the connected components in the graph and draw edges (i.e., induced pairs) between all vertices in a component. Again to ensure quality, before computing the induced pairs, we use the max-flow min-cut theorem \citep{dantzig1955max} considering the similarity scores as edge weights to limit the size of each component to 50 vertices (since ideally, a component should have at most 45 vertices, one summary from each language) and set their minimum acceptance threshold to $\tau' \leftarrow{} \tau - 0.10$.

We finally assembled the originally aligned pairs and induced pairs to create the CrossSum dataset. Figure \ref{fig:bubble} (Appendix) shows the article-summary statistics for all language pairs in CrossSum. As evident from the figure, CrossSum is not centered only around the English language but rather distributed across multiple languages.

\paragraph{Implicit Leakage} We initially made the train-dev-test splits respecting the original XL-Sum splits and performed an initial assessment of CrossSum by training a many-to-one model (articles written in any source language being summarized into one target language). Upon evaluation, we found very high ROUGE-2 scores (around 40) for many language pairs, even reaching as high as 60 for some (Figure \ref{fig:leak}). In contrast, \citet{hasan-etal-2021-xl} reported ROUGE-2 in the 10-20 range for the multilingual summarization task. 

We inspected the model outputs and found that many summaries were the same as the references. Through closer inspection, we found that their corresponding articles had a parallel counterpart occurring in the training set in some other language. During training, the model was able to align the representations of parallel articles (albeit written in different languages) and generate the same output by memorizing from the training sample. While models should undoubtedly be credited for being able to make these cross-lingual mappings, this is not ideal for benchmarking purposes as this creates unusually high ROUGE scores. We denote this phenomenon as `\textit{implicit leakage}' and make a new dataset split to avoid this. Before proceeding, we deduplicate the XL-Sum dataset\footnote{XL-Sum has been deduplicated using lexical overlap methods only. But due to the risk of implicit leakage, which is not lexical, we further perform semantic deduplication.} using semantic similarity, considering two summaries $S_A, S_A'$ in language \texttt{A} to be duplicates of one another if their LaBSE representations have similarity above $0.95$. We take advantage of the component graph mentioned previously to address the leakage and assign all article-summary pairs originating from a single component in the training (dev/test) set of CrossSum, creating an 80\%-10\%-10\% split for all language pairs. Since parallel articles no longer appear in the training set of one and the dev/test set of another, the leakage is not observed anymore (Figure \ref{fig:leak}). We further validated this by inspecting the model outputs and found no exact copies. 

\section{Human Evaluation of CrossSum}\label{sec:human_eval}

To establish the validity of our automatic alignment pipeline, we conducted a human evaluation to study the quality of the cross-lingual alignments. 

We selected all possible combinations of language pairs from a list of nine languages ranging from high-resource to low-resource to assess the alignment quality in different pair configurations (e.g., high-high, low-high, low-low) as per the language diversity categorization by \citet{joshi-etal-2020-state}. We chose three high-resource languages, English, Arabic, and (simplified) Chinese (categories 4 and 5); three mid-resource languages, Indonesian, Bengali, and Urdu (category 3); and three low-resource languages, Punjabi, Swahili, and Pashto (categories 1 and 2), as representative languages and randomly sampled fifty cross-lingual summary alignments from each language pair for annotation. As a direct evaluation of these pairs would require bilingually-proficient annotators for both languages, which are practically intractable for distantly related languages (e.g., Bengali-Swahili), we resorted to a pivoting approach during annotation for language pairs that do not contain English. For a language pair $(l_1-l_2)$, where $l_1 \neq en$ and $l_2 \neq en$, we sampled alignments $(x, y)$ such that $\exists (x, e) \in (l_1-en)$ and $\exists (y, e) \in (l_2-en)$, for an English article $e$.  In other words, we ensure that both the articles of the sampled cross-lingual pair have a corresponding cross-lingual pair with an English article. An alignment $(x, y)$ would be deemed correct if both $(x, e)$ and $(y, e)$ are correct. This formulation thus reduced the original problem to annotating samples from language pairs $(l_1-en)$ and $(l_2-en)$, where $l_1$ and $l_2$ are from the previously selected languages that are not English. 

We hired bilingually proficient expert annotators adept in the language of interest and English. Two annotators labeled each language pair where one language is English. We presented them with corresponding summaries of the cross-lingual pairs (and optionally the articles themselves) and elicited yes/no answers to the question: \par\textit{``Can the provided sequences be considered summaries for the same article?''}\footnote{We do not explicitly evaluate article-summary correctness as this has already been studied in work on XL-Sum. This was also done to reduce annotation costs.}

\begin{figure}[t]
\centering
\includegraphics[width=0.5\textwidth]{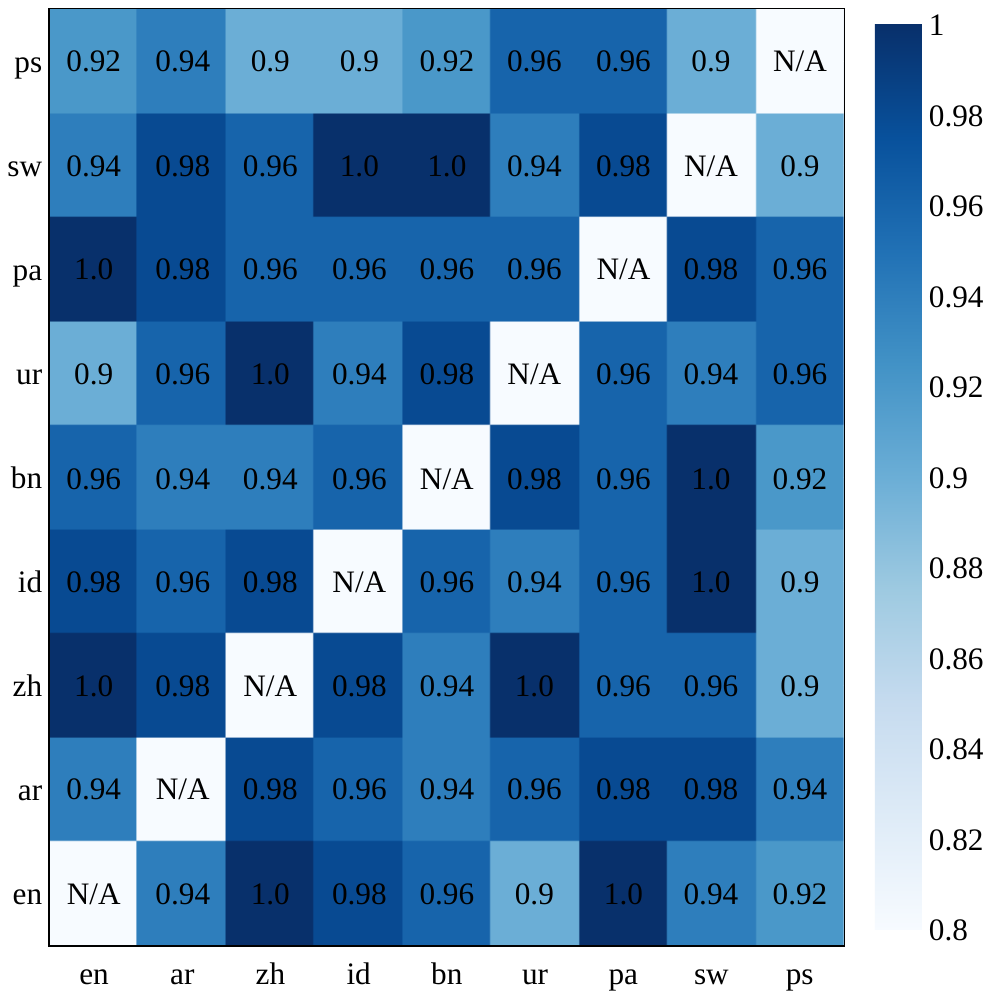}
\caption{A heatmap showing alignment accuracies of different language pairs obtained by human evaluation.}
\label{fig:human_eval_heatmap}
\end{figure}

We deem a sequence pair accurate if both annotators judge it as valid. We show the alignment accuracies of the language pairs in Figure \ref{fig:human_eval_heatmap}. 

As evident from the figure, the annotators judge the aligned summaries to be highly accurate, with an average accuracy of 95.67\%. We used Cohen's Kappa \citep{cohen1960coefficient} to establish the inter-annotator agreement and show the corresponding statistics in Table \ref{tab:iaa-kappa} in the Appendix.

\section{Training \& Evaluation Methodologies}\label{sec:traineval}

In this section, we discuss the multistage sampling strategy for training cross-lingual text generation models and our proposed metric for evaluating model-generated summaries.
\subsection{Multistage Language Sampling (MLS)}

From Figure \ref{fig:bubble}, it can be observed that CrossSum is heavily imbalanced. Thus, training directly without upsampling low-resource languages may result in their degraded performance. \citet{conneau2019unsupervised} used probability smoothing for upsampling in multilingual pretraining and sampled all examples of a batch from one language. However, extending this technique to the language pairs in CrossSum would result in many batches having repeated samples as many language pairs do not have enough training samples in total compared to the batch sizes used in practice (e.g., \citet{conneau2019unsupervised} used a batch size of 256, which exceeds the training set size of nearly 1,000 language pairs in CrossSum). At the same time, many language pairs would not be sampled during training for lack of enough training steps (due to our constraints on computational resources). To address this, we adapt their method to introduce a \textbf{M}ultistage \textbf{L}anguage \textbf{S}ampling algorithm (\textbf{MLS}) to ensure that the target summaries of a batch are sampled from the same language.

Let $L_1, L_2, \ldots, L_n$ be the languages of a cross-lingual source-target dataset, and $c_{ij}$ be the number of training samples where the target is from $L_i$ and source from $L_j$. We compute the probability $p_i$ of each target language $L_i$ by 
\begin{equation*}
    p_i = \frac{\sum_{k=1}^{n}c_{ik}}{\sum_{j=1}^{n}\sum_{k=1}^{n}c_{jk}} \quad \forall i \in \{1, 2, \ldots, n\}
\end{equation*}

We then use an exponent smoothing factor $\alpha$ and normalize the probabilities 
\begin{equation*}
    q_i =\frac{p_i^\alpha}{\sum_{j=1}^{n} p_j^\alpha} \quad\forall i \in \{1, 2, \ldots, n\}
\end{equation*}

Given the target language $L_i$, we now compute the probability of a source language $L_j$, represented by $p_{j|i}$. 
\begin{equation*}
    p_{j|i} = \frac{c_{ij}}{\sum_{k=1}^{n}c_{ik}} \forall j \in \{1, 2, \ldots, n\}
\end{equation*}

We again smooth $p_{j|i}$ by a factor $\beta$ and obtain the normalized probabilities \begin{equation*}
    q_{j|i} = \frac{p_{j|i}^\beta}{\sum_{k=1}^{n}p_{k|i}^{\beta}} \forall j \in \{1, 2, \ldots, n\}
\end{equation*}

Using the probabilities, we describe the training process with the MLS algorithm in Algorithm \ref{alg:samp}. 

\begin{algorithm}[t]
\DontPrintSemicolon
    \vspace{1mm}
    \KwInput{$D_{ij}\ \forall i,j \in \{1, 2, \ldots, n\}$: training data with tgt/src languages $L_i$/$L_j$; $c_{ij} \leftarrow |D_{ij}|\ \forall i,j \in \{1, 2, \ldots, n\}$; $m$: number of mini-batches.}\vspace{1mm}
    \hrule
    \vspace{1mm}
    Compute $q_i, q_{j|i}$ using $c_{ij}$\\
    \While{(\texttt{Model Not Converged})}{
        $batch \leftarrow \phi$\\
        Sample $L_i \sim q_i$\\
        \For{$k \leftarrow 1$ \KwTo $m$}{
            Sample $L_j \sim q_{j|i}$\\
            Create mini-batch $mb$ from $D_{ij}$\\
            $batch \leftarrow batch \cup \{mb\}$
        }
        Update model parameters using $batch$%
    }
\caption{Multistage Language Sampling (MLS)}
\label{alg:samp}
\end{algorithm}

Note that the proposed algorithm can be applied to any cross-lingual seq2seq task where both the source and target languages are imbalanced.

\subsection{Evaluating Summaries Across Languages}\label{sec:lase}

A sufficient number of reference samples are essential for the reliable evaluation of model-generated summaries. However, for many CrossSum language pairs, even the training sets are small, let alone the test sets (the median size is only 33). For instance, the Japanese-Bengali language pair has 34 test samples only, which is too few for reliable evaluation. But the size of the in-language\footnote{Both article and summary belonging to the same language} test sets of Japanese and Bengali are nearly 1,000. Being able to evaluate against reference summaries written in the source language would thus alleviate this insufficiency problem by leveraging the in-language test set of the source language. 

For this purpose, cross-lingual similarity metrics that do not rely on lexical overlap (i.e., unlike ROUGE) are required. Embedding-based similarity metrics \citep{Zhang2020BERTScore,zhao-etal-2019-moverscore} have recently gained popularity. We draw inspiration from them and design a similarity metric that can effectively measure similarity across languages in a language-independent manner. We consider three essential factors:

\noindent\textbf{1. Meaning Similarity}: The generated and reference summaries should convey the same meaning irrespective of their languages. Just like our alignment procedure from Section \ref{sec:cross}, we use LaBSE to compute the meaning similarity between the generated ($s_{gen}$) and reference summary ($s_{ref}$):
    \begin{equation*}
        \text{MS}(s_{gen}, s_{ref}) = \text{emb}(s_{gen})^{\text{T}} \text{emb}(s_{ref})
    \end{equation*}
    where $\text{emb}(s)$ denotes the embedding vector output of LaBSE for input text $s$. 
    
\noindent\textbf{2. Language Confidence}: The metric should identify, with high confidence, that the summary is indeed being generated in the target language. As such, we use the \emph{fastText} language-ID classifier \citep{joulin-etal-2017-bag} to obtain the language probability distribution of the generated summary and define the Language Confidence (LC) as:
    \begin{equation*}
    \text{LC}(s_{gen}, s_{ref}) = \begin{cases}
        1 \text{, if } L_{ref} = \argmax P(L_{gen})\\
        P(L_{gen}=L_{ref}) \text{, otherwise}\\
    \end{cases}
    \end{equation*}

\noindent\textbf{3. Length Penalty}: Generated summaries should not be unnecessarily long, and the metric should penalize long summaries. While model-based metrics may indicate how similar a generated summary is to its reference and language, it is unclear how they can be used to determine its brevity. As such, we adapt the BLEU \citep{papineni2002bleu} brevity penalty to measure the length penalty:
    \begin{equation*}
    \text{LP}(s_{gen}, s_{ref}) = \begin{cases}
        1 \text{, if } |s_{gen}| \leq |s_{ref}| + c\\
        \exp(1 - \frac{|s_{gen}|}{|s_{ref}| + c}) \text{, otherwise}\\
    \end{cases}
    \end{equation*}

$s_{gen}$ and $s_{ref}$ may not be of the same language, and parallel texts may vary in length across languages. Hence, we use a length offset $c$ to avoid penalizing generated summaries slightly longer than the references. By examining the standard deviation of mean summary lengths of the languages, we set $c = 6$.
    
We finally define our metric, \textbf{L}anguage-\textbf{a}gnostic \textbf{S}ummary \textbf{E}valuation (\textbf{LaSE}) score as follows.
\begin{multline*}
\text{LaSE}(s_{gen}, s_{ref}) = \text{MS}(s_{gen}, s_{ref}) \\
\times \text{LC}(s_{gen}, s_{ref}) \times \text{LP}(s_{gen}, s_{ref})
\end{multline*} 

\begin{figure*}[!tbh]
\centering
\includegraphics[width=\textwidth]{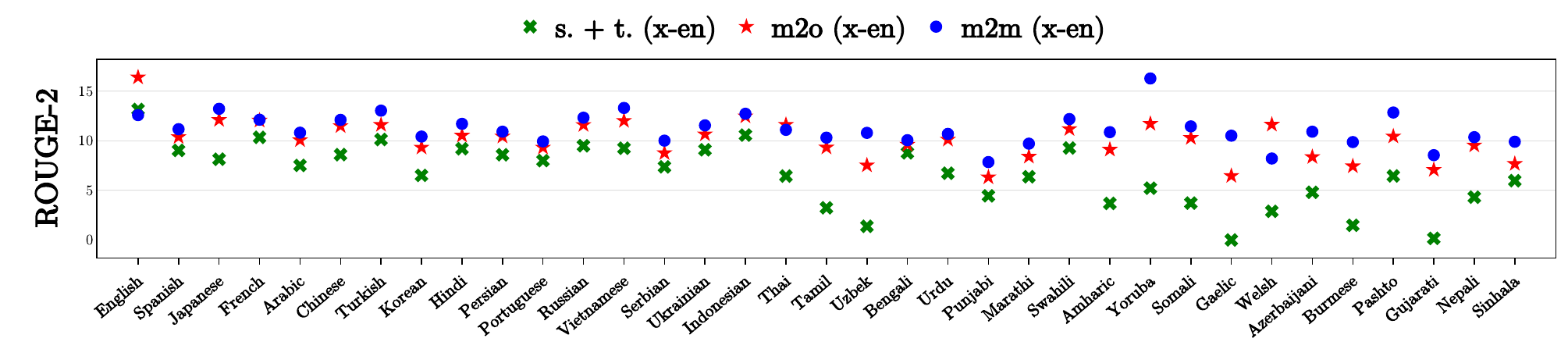}
\includegraphics[width=\textwidth]{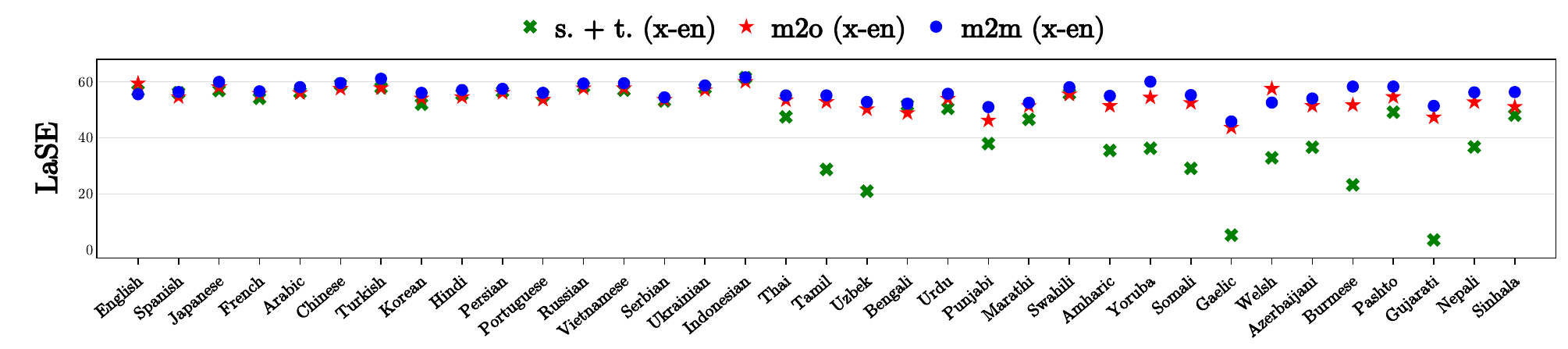}
\includegraphics[width=\textwidth]{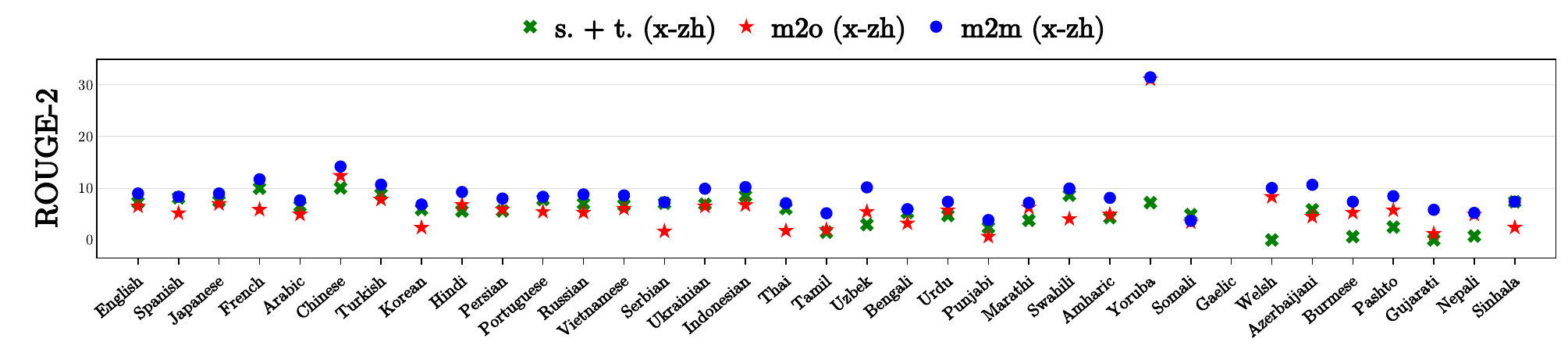}
\includegraphics[width=\textwidth]{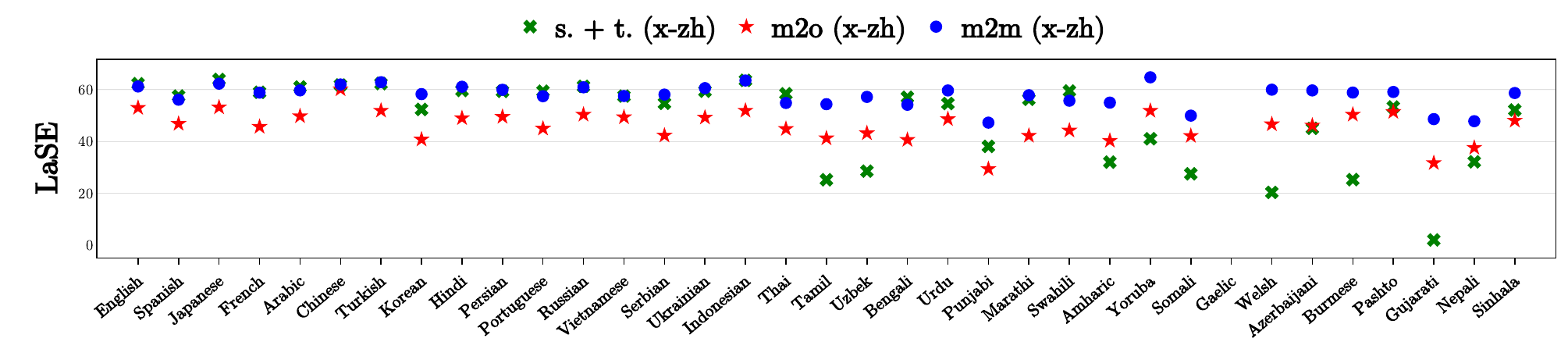}
\caption{ROUGE-2 and LaSE scores for English and Chinese as target languages as the source languages vary. The m2m model significantly outperforms the m2o models and summarize-then-translate baseline in most languages. The comparisons with other target languages are shown in the Appendix (Figure \ref{fig:m2o-app}) due to space limitations.}
\label{fig:m2o}
\end{figure*}

\begin{figure*}[!tbh]
\centering
\includegraphics[width=\textwidth]{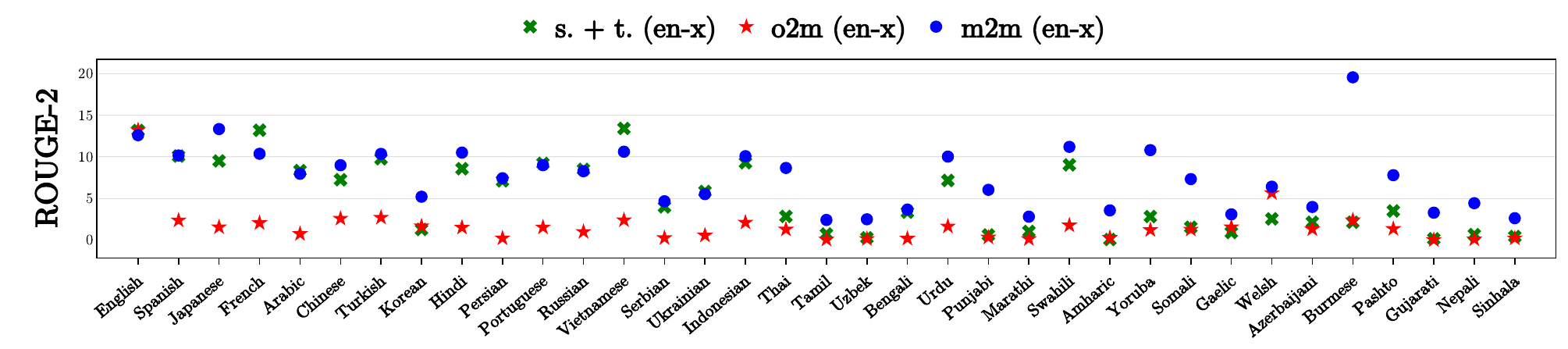}
\includegraphics[width=\textwidth]{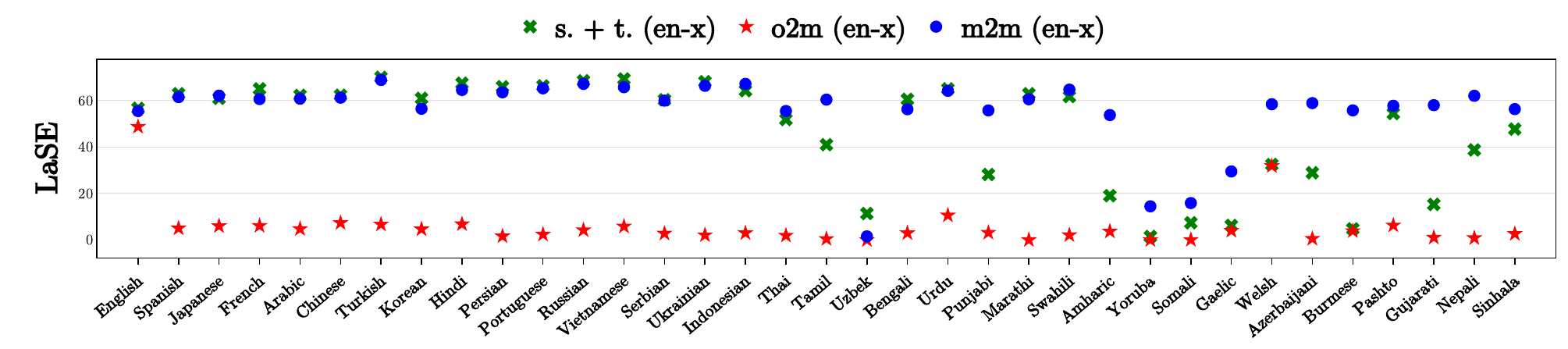}
\includegraphics[width=\textwidth]{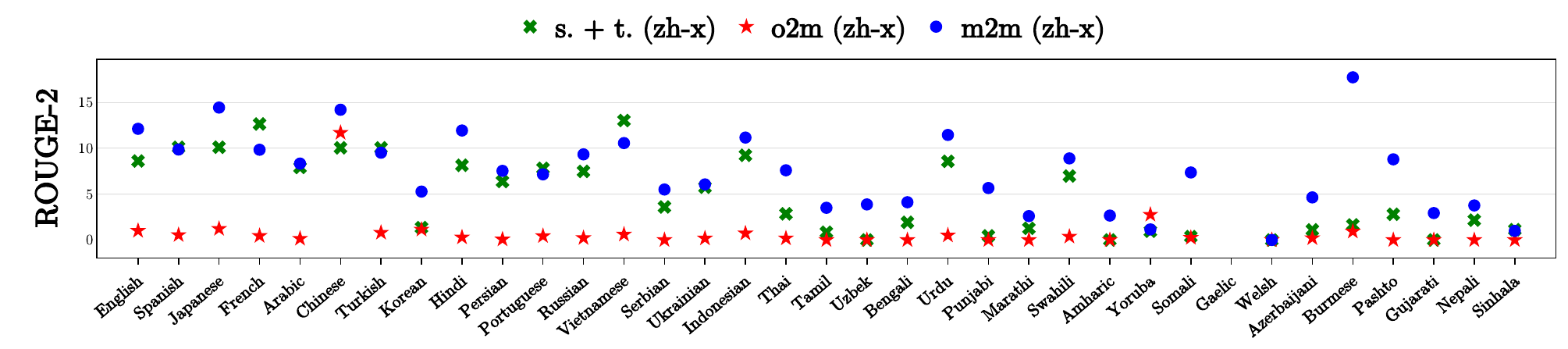}
\includegraphics[width=\textwidth]{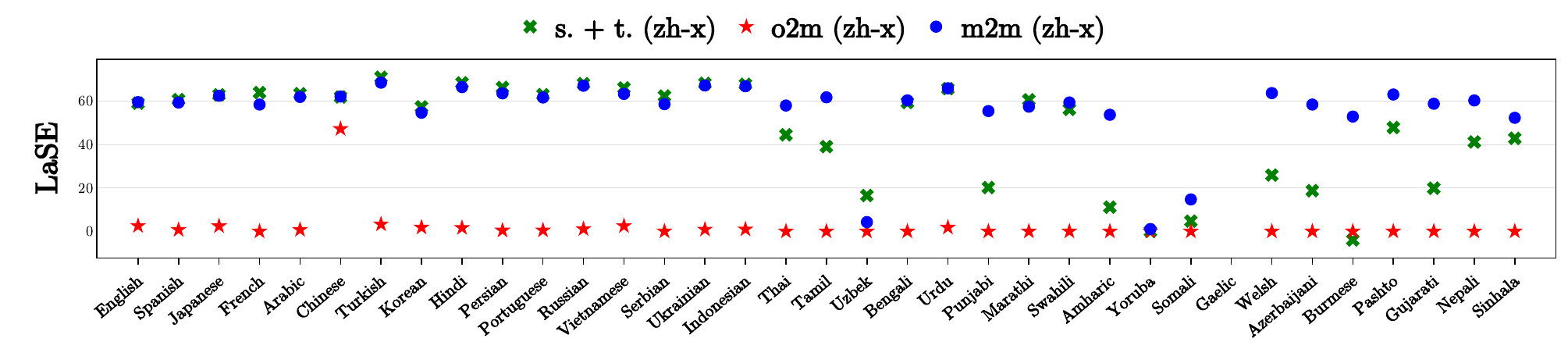}
\caption{ROUGE-2 and LaSE scores for English and Chinese as source languages as the target languages vary. The m2m model significantly outperforms the o2m models and summarize-then-translate baseline in most languages. The comparisons with other source languages are shown in the Appendix (Figure \ref{fig:o2m-app}) due to space limitations.}
\label{fig:o2m}
\end{figure*}

\section{Experiments \& Discussions}\label{sec:exp}

One model capable of generating summaries in any target language for an input article from any source language is highly desirable. However, it may not be the case that such a `many-to-many' model (m2m in brief) would outperform many-to-one (m2o) or one-to-many (o2m) models\footnote{Discussed in detail in Appendix \ref{sec:modeling_details}.}, which are widely-used practices for XLS \citep{ladhak-etal-2020-wikilingua, perez-beltrachini-lapata-2021-models}. In this section, we establish that the m2m model, trained in the presence of samples from all possible language pairs using the MLS algorithm from Section \ref{sec:traineval}, consistently outperforms m2o, o2m, and summarize-then-translate (s.+t.) baselines given equal training steps.

In addition to the proposed m2m model, we train five different m2o and o2m models using five highly spoken\footnote{\url{https://w.wiki/Pss}} and typologically diverse pivot (i.e., the `one' in m2o and o2m) languages: English, Chinese (simplified), Hindi, Arabic, and Russian. As another baseline, we use a summarize-then-translate pipeline. As fine-tuning pretrained language models \citep{devlin2018bert, xue-etal-2021-mt5} have shown state-of-the-art results on monolingual and multilingual text summarization \citep{rothe-etal-2020-leveraging, hasan-etal-2021-xl}, we fine-tune each model using a pretrained mT5 \citep{xue-etal-2021-mt5} by providing explicit cross-lingual supervision. We show the results on ROUGE-2 F1 and LaSE in Figures \ref{fig:m2o} and \ref{fig:o2m}\footnote{A detailed description of the training procedures and hyperparameter choices are detailed in Appendix \ref{sec:training_setups}.}. We limit our evaluation only to the languages supported by mT5, fastText, and M2M-100 (the translation model used in s.+t.). 

Results indicate that the m2m model consistently outperforms m2o, o2m, and s.+t., with an average ROUGE-2 (LaSE) score of 8.15 (57.15) over all languages tested, 3.12 (9.02) above s.+t. Moreover, compared to the o2m models on language pairs where the pivots are the targets, the m2m model scores 1.80 (5.84) over m2os, and on those where the pivots are the sources, 6.52 (51.80) over o2ms.

Upon inspection of the model outputs, we found the m2o models to be able to generate non-trivial summaries. In contrast, the o2m models completely failed to produce cross-lingual summaries, performing in-language summarization (the language of the summary is the same as that of its input article) for all targets. We hypothesize that varying the target language in a batch hampers the decoder's ability to generate from a specific language, possibly because of the vast diversity of target languages in the batch (discussed further in Appendix \ref{sec:ablation})\label{foot:}. s.+t. performed well on high-resource languages but poorly on low-resource ones. This was revealed to be a limitation of the translation model used in the pipeline. 

\subsection{Zero-shot Cross-lingual Transfer}\label{sec:zero_shot}

The previous experiments were done in a fully supervised fashion. However, for many low-resource language pairs, samples are not abundantly available. Hence, it is attractive to be able to perform zero-shot cross-lingual generation \citep{duan-etal-2019-zero} without relying on any labeled examples. 

To this end, we fine-tuned mT5 with only the in-language samples (i.e., the source and target both have the same language) in a multilingual fashion and, during inference, varied the target language. Unfortunately, the model totally fails at generating cross-lingual summaries and performs in-language summarization instead. 

We also fine-tuned m2o models (with only the in-language samples of the target language) in a monolingual fashion and ran inference in a zero-shot setting with samples from other languages as input. Here, the models are able to generate non-trivial summaries for some language pairs but still lag behind fully supervised models by a significant margin. We have included Figures \ref{fig:m2o-zs} and \ref{fig:o2m-zs} in the Appendix to illustrate this. 

Furthermore, we ran inference with the m2m model on distant low-resource language pairs that were absent in training. Their LaSE scores were substantially below supervised pairs, meaning zero-shot transfer in supervised multilingual models \citep{johnson2017google} shows weak performance. 

We do not perform few-shot experiments and leave them as potential future directions.

\section{Analysis of Results}\label{sec:significance} 

\paragraph{Statistical significance} While the scores obtained from the experiments in Section \ref{sec:exp} indicate that the proposed m2m model performs better than the others, the differences are very close in many language pairs. Therefore, a statistical significance test is still warranted to support our claim further. As such, for each language pair experimented on, we performed the Bootstrap resampling test \citep{koehn-2004-statistical} with the m2m model against the best-performing model among the others in a one vs. all manner: if m2m has the best (ROUGE-2/LaSE) score, we compare it with the model with the second-best score, and if m2m is not the best, we compare it with the best. 

\begin{table}[!thp]
\centering\setlength{\tabcolsep}{2pt}
{%
\begin{tabular}{ccccc}
\hline
Pivot & Metric & Better & Worse & Insignificant\\
\hline
x-en & R-2/LaSE & 8/18 & 2/2 & 25/15\\
en-x & R-2/LaSE & 20/15 & 3/14 & 12/6\\
x-zh & R-2/LaSE & 11/13 & 0/0 & 23/21\\
zh-x & R-2/LaSE & 17/12 & 1/2 & 16/20\\
x-hi & R-2/LaSE & 18/15 & 1/6 & 15/13\\
hi-x & R-2/LaSE & 19/15 & 0/6 & 15/13\\
x-ar & R-2/LaSE & 6/15 & 2/3 & 26/16\\
ar-x & R-2/LaSE & 23/15 & 1/5 & 10/14\\
x-ru & R-2/LaSE & 6/11 & 2/7 & 26/16\\
ru-x & R-2/LaSE & 19/13 & 2/7 & 13/14\\
\hline
\end{tabular}
}
\caption{Significance test on different pivot languages.}
\label{tab:significance}
\end{table}

Results ($p < 0.05$) in Table \ref{tab:significance} reveal that in more than 42\% language pairs tested, m2m is significantly better, and in less than 10\% pairs, it is considerably worse.\footnote{The numbers are even better if compared one vs. one. 
} This provides additional evidence in support of our claim that the m2m model performs better than others.

\paragraph{How reliable is LaSE?} At first, we validated the reliability of LaSE by showing its correlation with ROUGE-2. We took different checkpoints of the in-language summarization model used in s.+t. and computed ROUGE-2 and LaSE for the nine languages in Section \ref{sec:human_eval} for each checkpoint. The correlation coefficients of the calculated scores are shown in the second column of Table \ref{tab:correlation}. For all languages (from high- to low-resource), LaSE has a near-perfect correlation with ROUGE-2. 

However, the purpose of LaSE is to show that it is language-agnostic and can even be computed in the absence of references in the target language. Therefore, we evaluate the summaries with references in a different language from the target using the m2m model. For each target language, we first compute the standard LaSE for different source languages (denoted as LaSE-in-lang). We again compute LaSE after swapping the reference texts with the references in the language of the input text\footnote{Our curation method ensures that such summaries always exist in the corresponding test sets.} (denoted as LaSE-out-lang). We then show the correlation between the two variants of LaSE in the third column of Table \ref{tab:correlation}\footnote{Since many test sets of the language pairs from Section \ref{sec:human_eval} have too few samples for reliable evaluation (e.g., Punjabi-Pashto), for each target language, we use only the top-5 source languages by the number of their test set samples.} for each target language. Results show a substantial correlation between the two variants of LaSE for all languages.

From these two experiments, we can conclude that LaSE is an ideal metric for the evaluation of summarization systems and can be computed in a language-independent manner.

\begin{table}[H]
\centering\setlength{\tabcolsep}{1pt}
\begin{tabular}{l c c}
\hline
Target & ROUGE-2 vs. & LaSE-in-lang vs.\\
Lang. & LaSE-in-lang. & LaSE-out-lang.\\
& Pearson/Spearman & Pearson/Spearman\\
\hline
English & 0.976/0.939 & 0.993/1.000\\
Arabic & 0.903/0.987 & 0.968/0.942\\
Chinese & 0.983/1.000 & 0.996/1.000\\
Indonesian & 0.992/0.975 & 0.872/0.828\\
Bengali & 0.947/0.902 & 0.819/0.771\\
Urdu & 0.997/0.951 & 0.774/0.828\\
Punjabi & 0.988/0.963 & 0.881/0.885\\
Swahili & 0.990/0.951 & 0.979/0.885\\
Pashto & 0.994/0.987 & 0.883/0.885\\
\hline
\end{tabular}
\caption{Correlation analysis of ROUGE-2 and LaSE. We compute both Pearson and Spearman coefficients.}
\label{tab:correlation}
\end{table}

\section{Related Works}

Pipeline-based methods were popular at the beginning stages of XLS research \citep{leuski2003cross, ORASAN08.539, wan-etal-2010-cross}, breaking the task into a sequence of summarization and translation tasks. End-to-end methods that performed XLS with a single model gained popularity with the emergence of neural models. \citet{8370729} used knowledge distillation \citep{44873} to train a student XLS model from two summarization and translation teacher models. Using a synthetic dataset, \citet{zhu-etal-2019-ncls, cao-etal-2020-jointly} performed XLS with a dual Transformer \citep{vaswani2017attention} architecture in a multitask framework, while \citet{bai-etal-2021-cross} proposed a single encoder-decoder for better transfer across tasks. \citet{chi-etal-2021-mt6} introduced multiple pretraining objectives specifically tailored to cross-lingual tasks that showed improved results on XLS. We refer our readers to \citet{10.1162/tacl_a_00520} for a more comprehensive literature review. 

Until recently, XLS was limited primarily to English-Chinese due to the lack of benchmark datasets. To promote the task beyond this language pair, \citet{ladhak-etal-2020-wikilingua} introduced Wikilingua, a large-scale many-to-one dataset with English as the pivot language, while \citet{perez-beltrachini-lapata-2021-models} introduced XWikis, containing 4 languages in 12 directions. 

More recently, \citet{wang2023zeroshot} explored zero-shot cross-lingual summarization by prompting \citep{10.1145/3560815} large language models like ChatGPT\footnote{https://openai.com/blog/chatgpt}, GPT-4 \citep{openai2023gpt4}, and BLOOMZ \citep{muennighoff2022crosslingual}.
\section{Conclusion \& Future Works}

In this work, we presented CrossSum, a large-scale, non-English-centric XLS dataset containing 1.68 million samples in 1,500+ language pairs. CrossSum provides the first publicly available XLS dataset for many of these pairs. Performing a limited-scale human evaluation of CrossSum, we introduced MLS, a multistage sampling algorithm for general-purpose cross-lingual generation, and LaSE, a language-agnostic metric for evaluating summaries when reference summaries in the target languages may not be available. We demonstrated that training one multilingual model can help towards better XLS than baselines. We also shed light on the potential to perform zero-shot and few-shot XLS with CrossSum. We share our findings and resources in the hopes of making the XLS research community more inclusive and diverse.

In the future, we will investigate the use of CrossSum for other summarization tasks, e.g., multi-document \citep{fabbri-etal-2019-multi} and multi-modal summarization \citep{zhu2018msmo}. We would also like to explore better techniques for m2m, zero-shot, and few-shot cross-lingual summarization.

\section*{Limitations}

Though we believe that our work has many merits, some of its limitations must be acknowledged. Despite exhaustive human annotation being the most reliable means of ensuring the maximum quality of a dataset, we had to resort to the automatic curation of CrossSum due to the enormous scale of the dataset. As identified in the human evaluation, not all of the alignments made by LaBSE are correct. They are primarily summaries describing similar (i.e., having a substantial degree of syntactic or semantic similarity) but non-identical events. LaBSE also fails to penalize numerical mismatches, especially if the summaries depict the same event. 

Consequently, any mistake made by LaBSE in the curation phase may propagate to the models trained using CrossSum. And since LaBSE is a component of the proposed LaSE metric, these biases may remain unidentified by LaSE in the evaluation stage. However, no matter which automatic method we use, there will be such frailties in these extreme cases. Since the objective of this paper is not to scrutinize the pitfalls of LaBSE but rather to use it as a means of curation and evaluation, we deem LaBSE the best choice due to its extensive language coverage and empirical performance in cross-lingual mining among existing alternatives.

\section*{Ethical Considerations}

\paragraph{License} CrossSum is a derivative of the XL-Sum dataset. XL-Sum has been released under the Creative Commons Attribution-NonCommercial-ShareAlike 4.0 International License (CC BY-NC-SA 4.0), allowing modifications and distributions for non-commercial research purposes. We are adhering to the terms of the license and releasing CrossSum under the same license.

\paragraph{Generated Text} All of our models use the mT5 model as the backbone, which is pretrained on a large multilingual text corpus. For a text generation model, even small amounts of offensive or harmful texts in pretraining could lead to dangerous biases in generated text \cite{luccioni-viviano-2021-whats}. Therefore, our models can potentially generate offensive or biased content learned during the pretraining phase, which is beyond our control. Text summarization systems have also been shown to generate unfaithful and factually incorrect (albeit fluent) \citep{maynez-etal-2020-faithfulness} texts. Thus, we suggest carefully examining the potential biases before considering them in any real-world deployment.

\paragraph{Human Evaluation} Annotators were hired from the graduates of an institute that provides professional training for many languages, including the ones evaluated in Section \ref{sec:human_eval}. Each annotator was given around 200-250 sequence pairs to evaluate. Each annotation took an average of one and a half minutes, with a total of approximately 5-6 hours for annotating the whole set. Annotators were paid hourly per the standard remuneration of bilingual professionals in local currency.

\paragraph{Environmental Impact} A total of 25 models were trained as part of this work. Each model was trained for about three days on a 4-GPU Tesla P100 server. Assuming 0.08 kg/kWh carbon emission\footnote{\url{https://blog.google/technology/ai/minimizing-carbon-footprint/}}, less than 175kg of carbon was released into the environment in this work, which is orders of magnitude below the most computationally demanding models.

\section*{Acknowledgements}

This work was funded by the Research and Innovation Centre for Science and Engineering (RISE), BUET. The OzSTAR national facility at Swinburne University of Technology was used to conduct the computational experiments. Funding for the OzSTAR program was provided in part by the Australian Government's Astronomy National Collaborative Research Infrastructure Strategy (NCRIS) allocation.


\bibliographystyle{style/acl_natbib}
\bibliography{anthology,emnlp2022}

\clearpage

\appendix 
\begin{large}
\noindent\textbf{Appendix}
\end{large}

\begin{figure*}[!tbh]
\centering
\includegraphics[width=\textwidth]{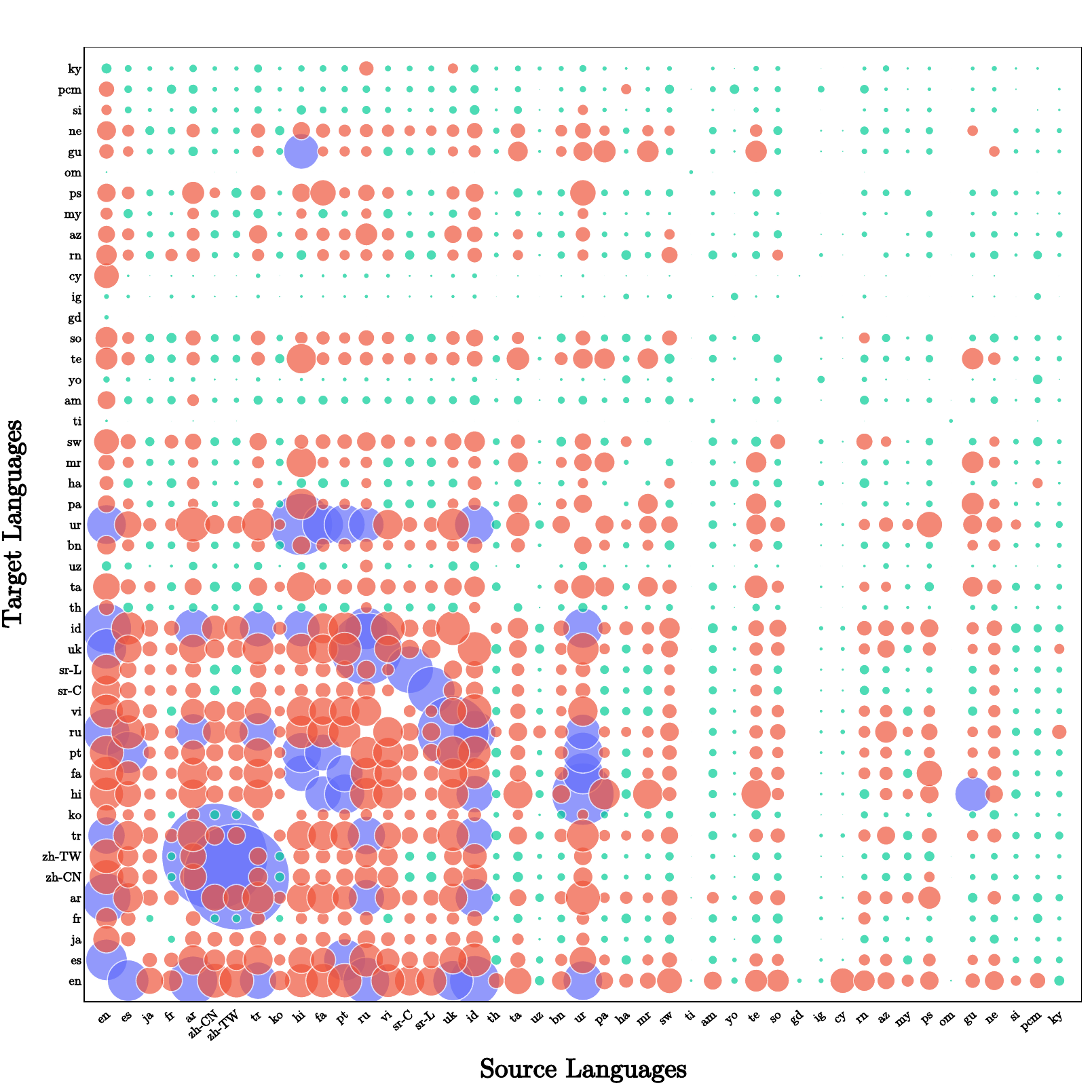}
\caption{A bubble plot depicting the article-summary frequencies of CrossSum. The radii of the bubbles are proportional to the number of samples for the corresponding language pair (exact numbers are in Table \ref{tab:crosssum}). Languages are ordered by the language taxonomy from \citet{joshi-etal-2020-state}. To show better contrast between language pairs, we color a bubble cyan if its frequency is below 500 (1218 pairs), red for 500 to 5000 (688 pairs), and blue for frequencies exceeding 5000 (52 pairs).}
\label{fig:bubble}
\end{figure*}

\section{Aligning Summaries using LaBSE}\label{sec:LaBSE}

In Section \ref{sec:cross}, we curated CrossSum by aligning parallel summaries in different languages. It might be argued why the articles themselves were not used for the alignment process. Initially, we experimented with whole-article embeddings. However, this resulted in many false-negative alignments, where similarity scores between parallel articles across languages were relatively low (verified manually between English and the authors' native languages). This is most likely attributed to the 512-token limit of LaBSE and different sequence lengths of those articles due to different languages having different subword segmentation fertility \citep{acs2019exploring}. This would entail that parallel articles in different languages might be truncated at different locations, resulting in discrepancies between their embeddings. As observed in the BUCC evaluation, LaBSE is well-suited for sentence-level retrieval. Since summaries are good representatives of entire articles, we finally chose summaries as our candidates for the alignment.

\section{Inter-annotator Agreement of Human Evaluation}

\begin{table}[H]
\centering\setlength{\tabcolsep}{2pt}
{%
\begin{tabular}{lc}
\hline
Language Pair & Cohen's Kappa\\
\hline
Arabic-English & 0.82\\
Chinese-English & 0.73\\
Indonesian-English & 0.73\\
Bengali-English & 0.73\\
Urdu-English & 0.76\\
Punjabi-English & 0.71\\
Swahili-English & 0.78\\
Pashto-English & 0.75\\
\hline
\end{tabular}
}
\caption{Language pair-wise kappa scores.}
\label{tab:iaa-kappa}
\end{table}

\begin{figure*}[!tbh]
\centering
\includegraphics[width=\textwidth]{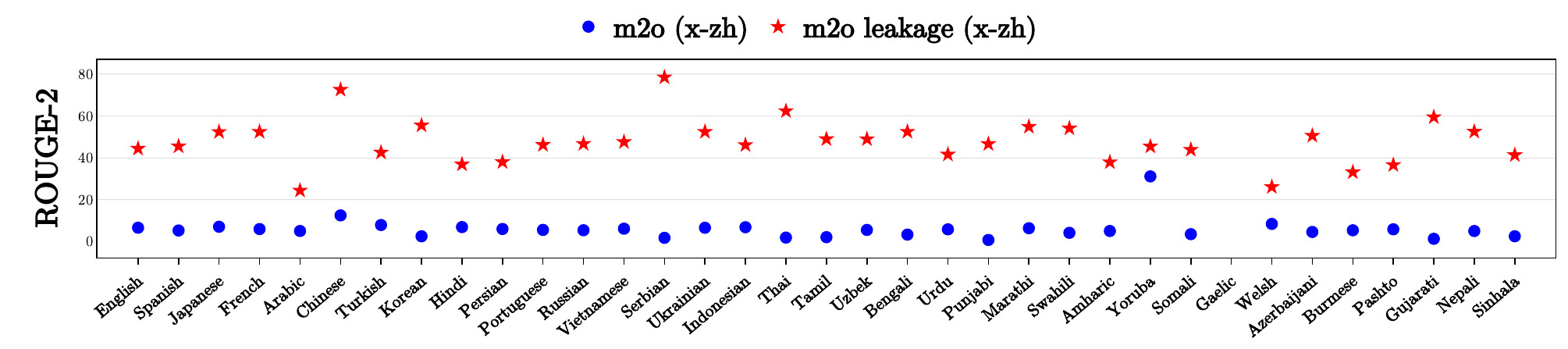}
\includegraphics[width=\textwidth]{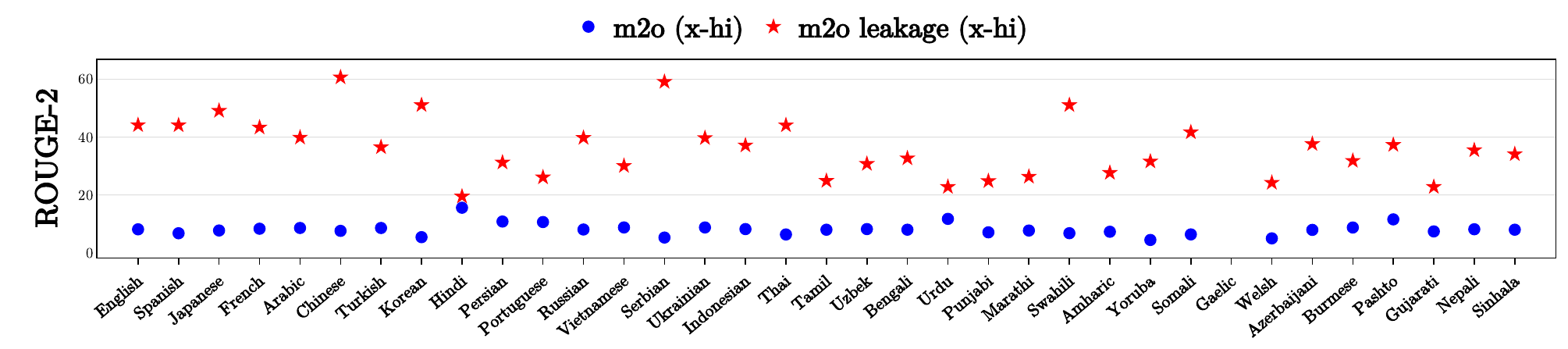}
\includegraphics[width=\textwidth]{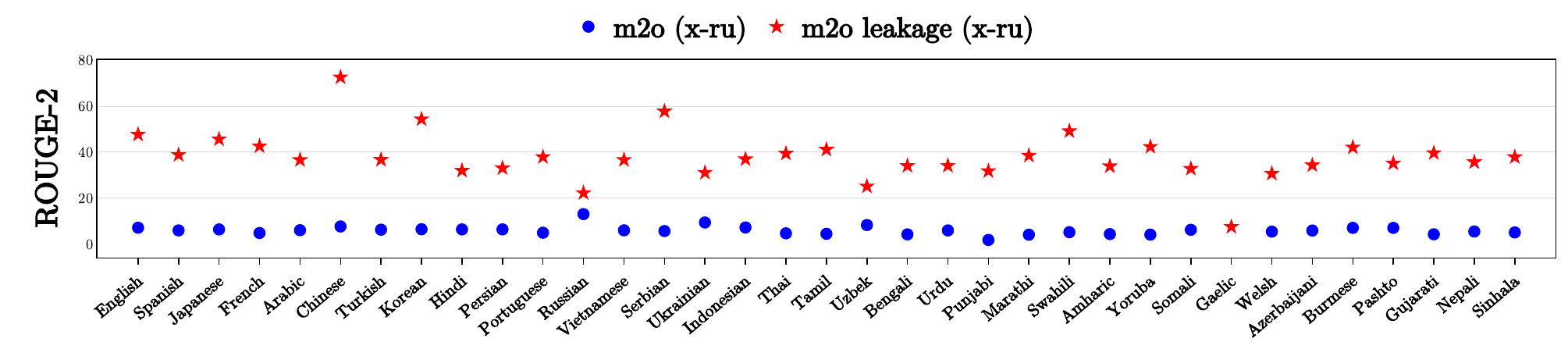}
\includegraphics[width=\textwidth]{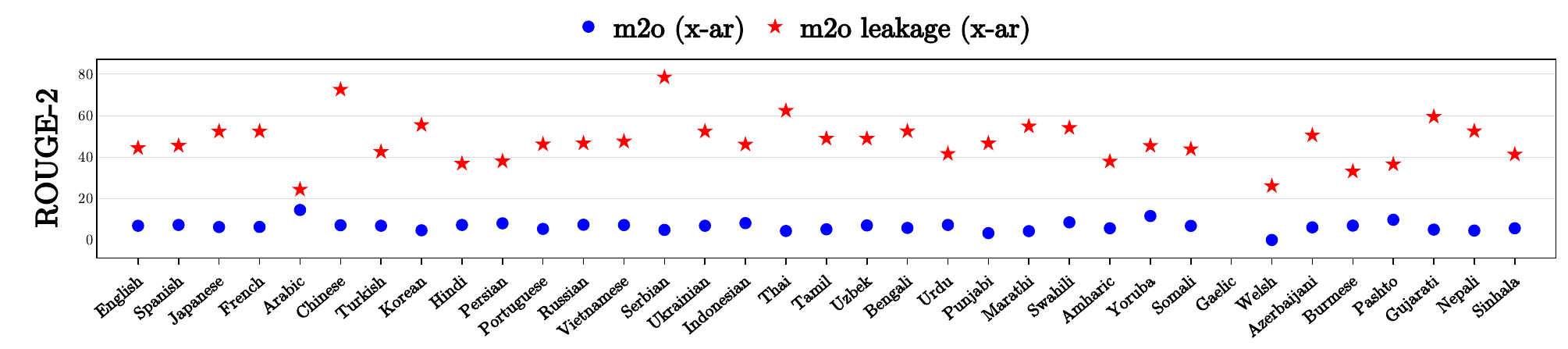}

\caption{Training on the dataset respecting the original XL-Sum splits causes absurdly high ROUGE scores (marked red) in many-to-one models due to implicit data leakage. Therefore, we split taking the issue into account, and consequently, models trained on the new set (marked blue) do not exhibit any unusual spike in ROUGE-2.}
\label{fig:leak-app}
\end{figure*}

\begin{figure*}[t]
\centering
\includegraphics[width=\textwidth]{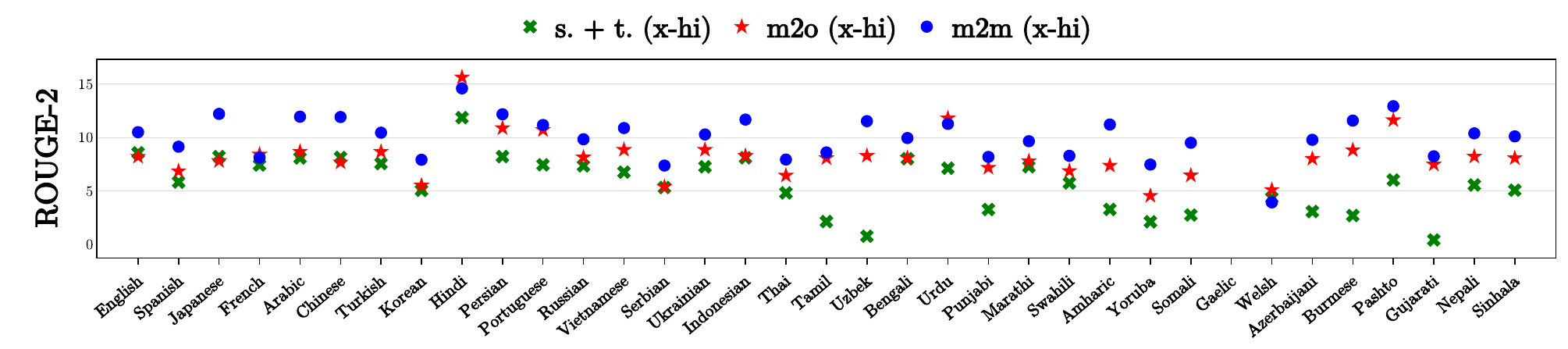}
\includegraphics[width=\textwidth]{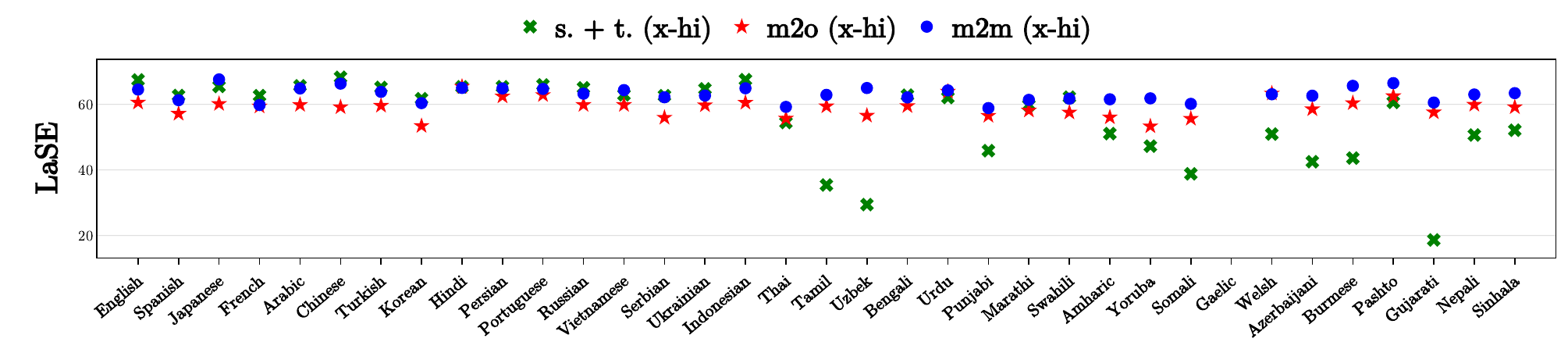}
\includegraphics[width=\textwidth]{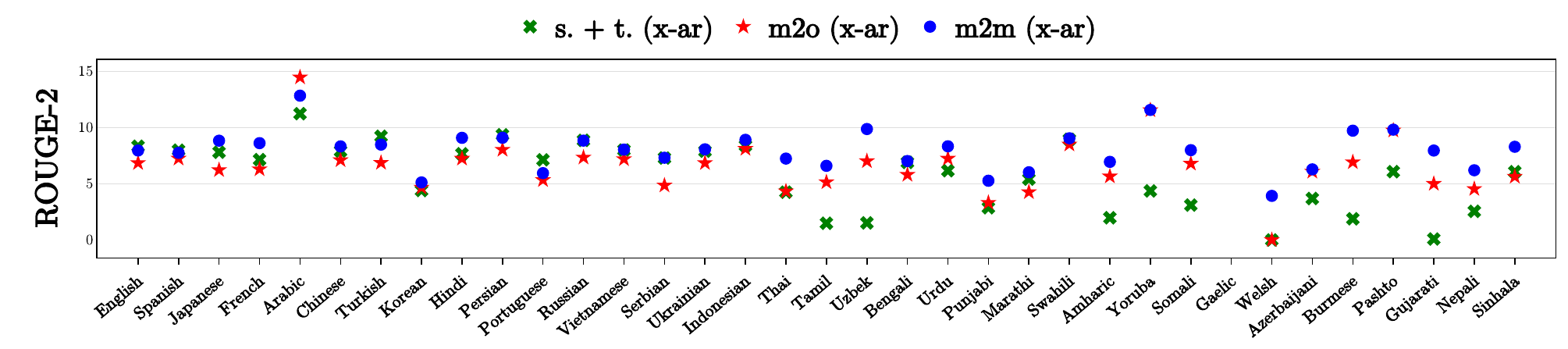}
\includegraphics[width=\textwidth]{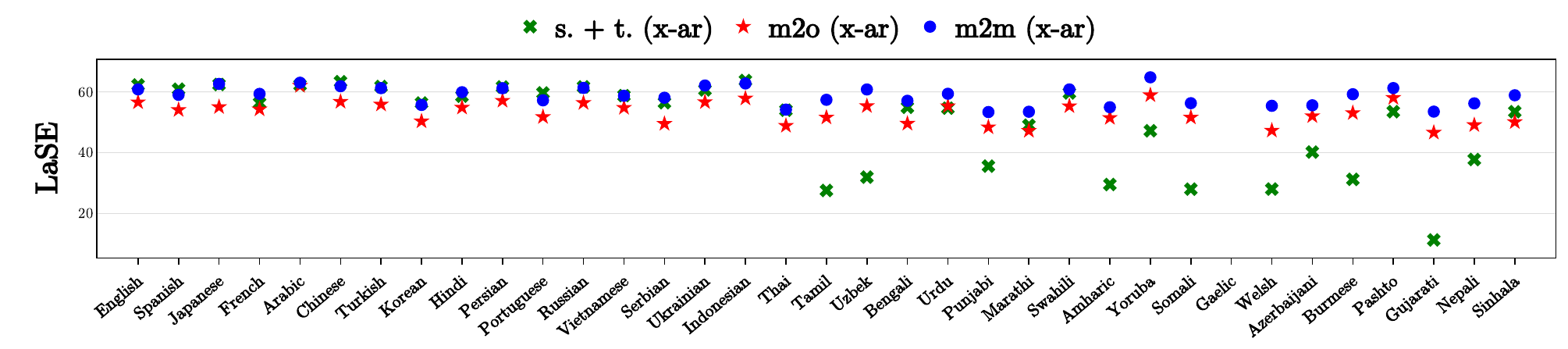}
\includegraphics[width=\textwidth]{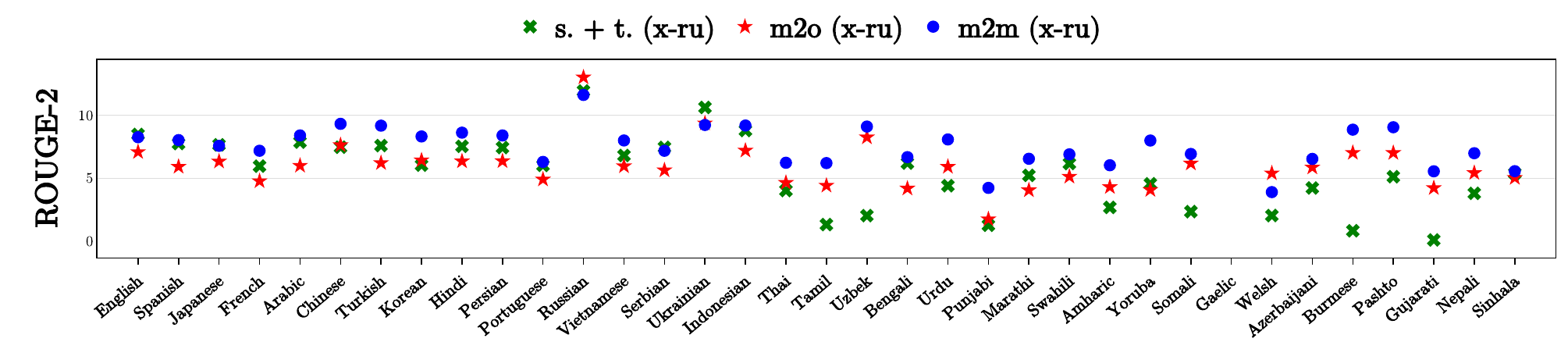}
\includegraphics[width=\textwidth]{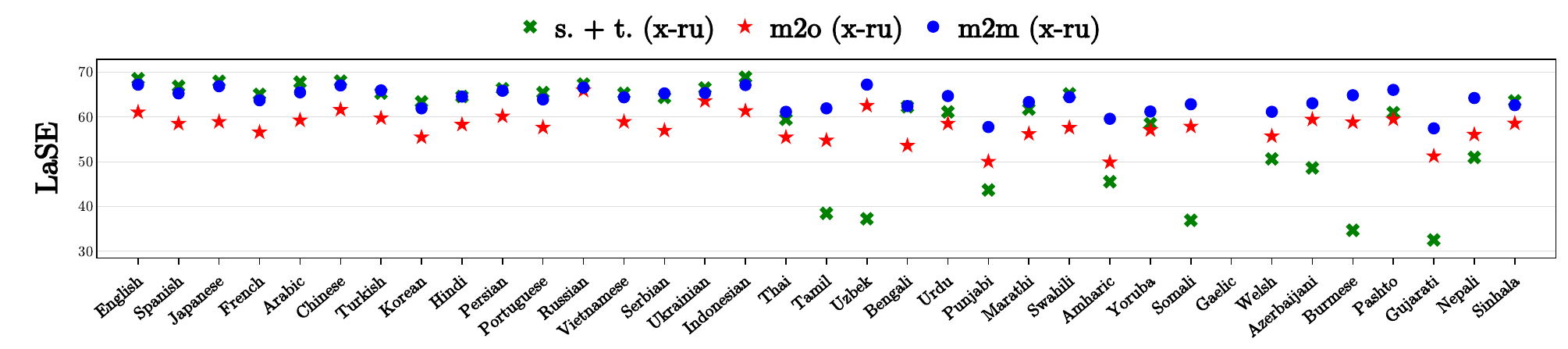}
\caption{ROUGE-2 and LaSE scores for Hindi, Arabic, and Russian as target pivots as the sources languages vary. Just like Figure \ref{fig:m2o}, the m2m model significantly outperforms the m2o models and s. + t. baseline on most languages.}
\label{fig:m2o-app}
\end{figure*}

\begin{figure*}[t]
\centering
\includegraphics[width=\textwidth]{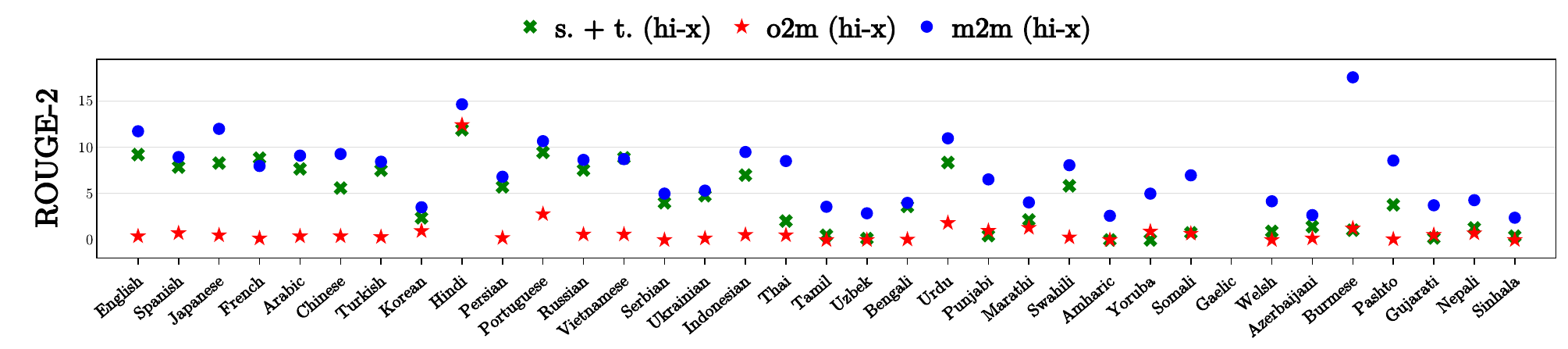}
\includegraphics[width=\textwidth]{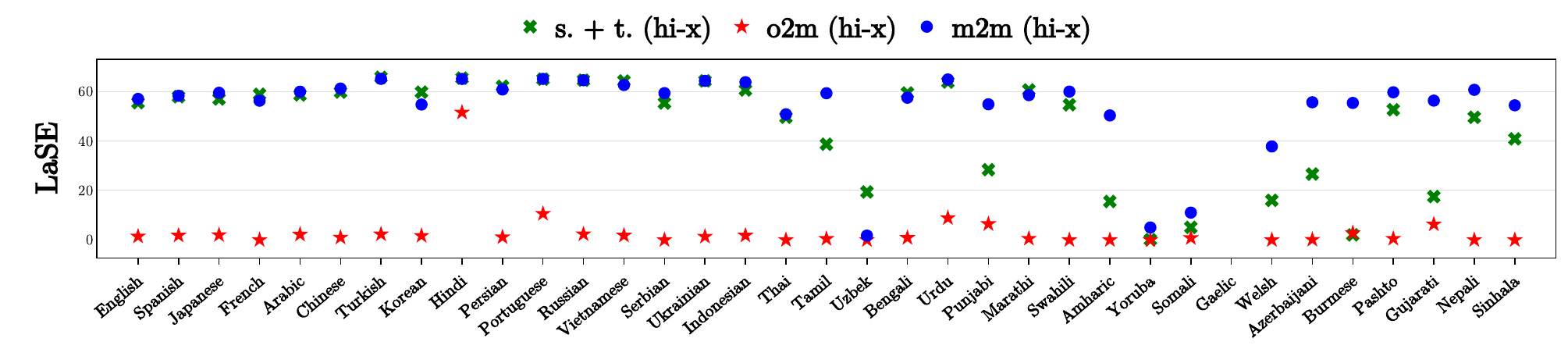}
\includegraphics[width=\textwidth]{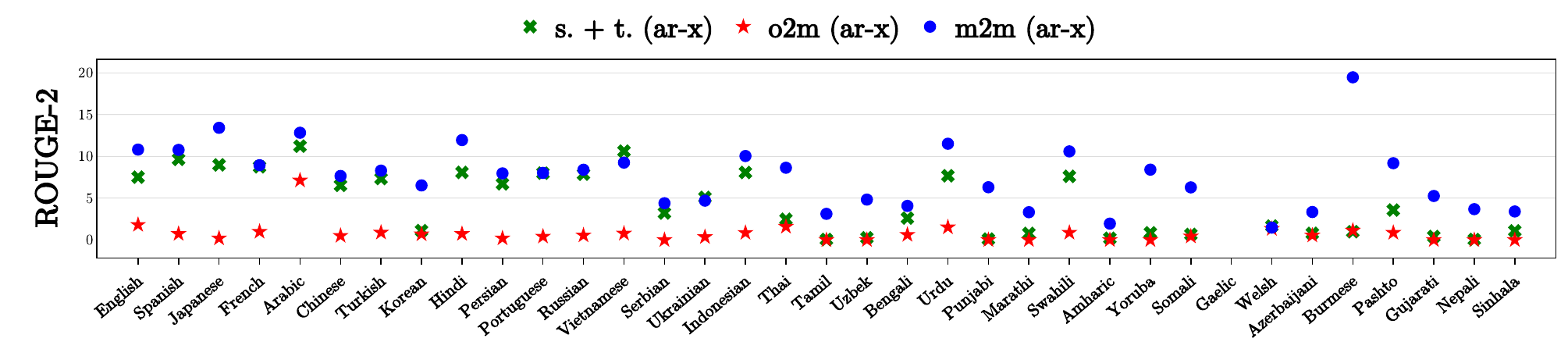}
\includegraphics[width=\textwidth]{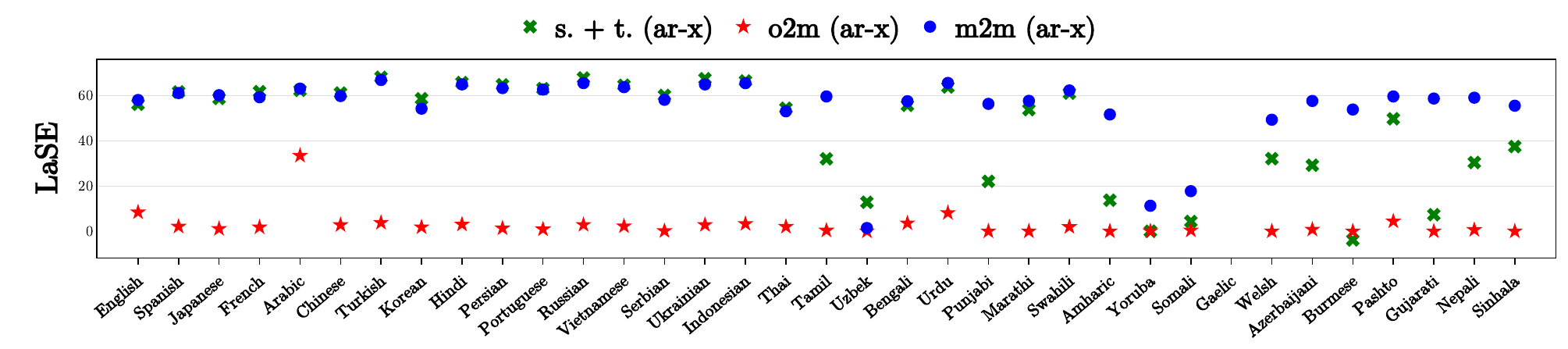}
\includegraphics[width=\textwidth]{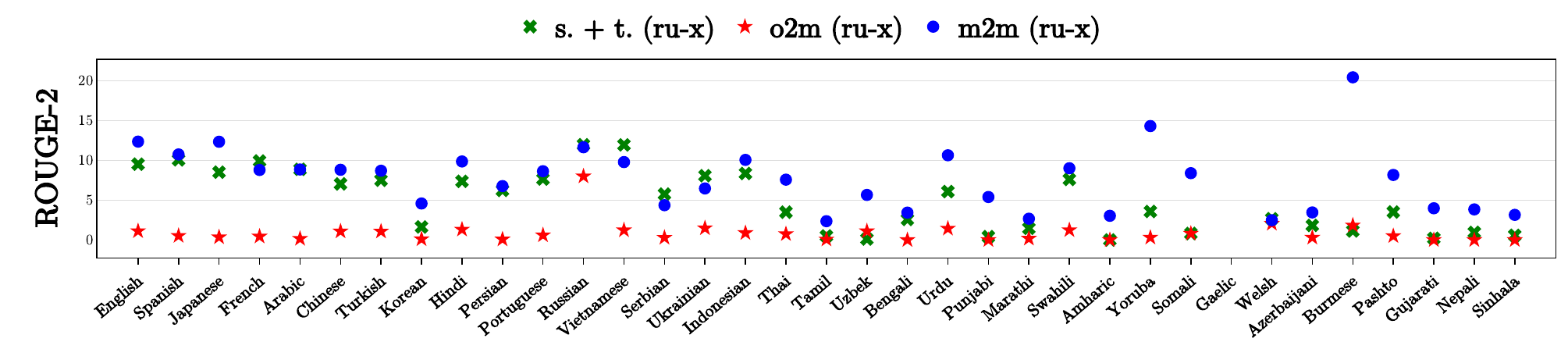}
\includegraphics[width=\textwidth]{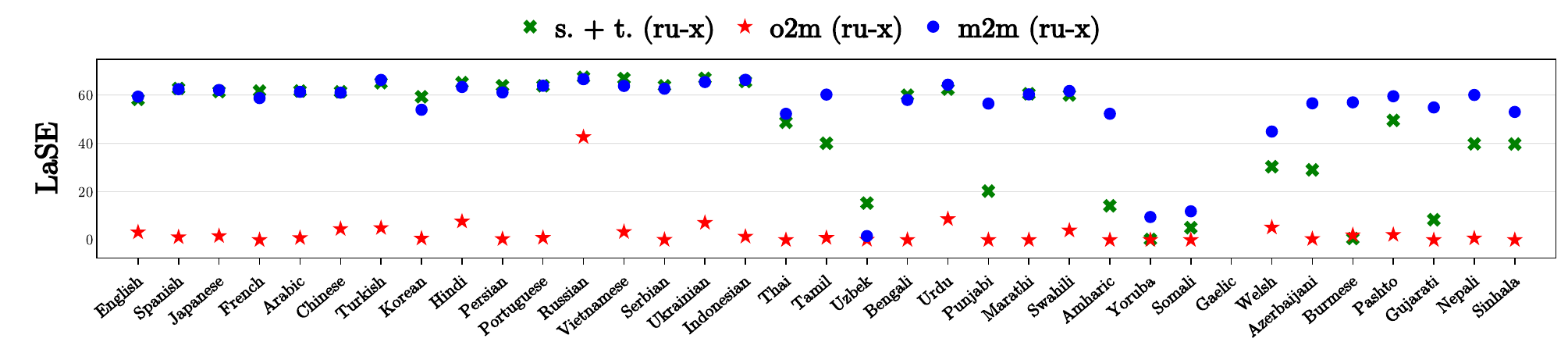}
\caption{ROUGE-2 and LaSE scores for Hindi, Arabic, and Russian as source pivots as the target languages vary. Just like Figure \ref{fig:o2m}, the m2m model significantly outperforms the o2m models and s. + t. baseline on most languages.}
\label{fig:o2m-app}
\end{figure*}

\begin{figure*}[t]
\centering
\includegraphics[width=\textwidth]{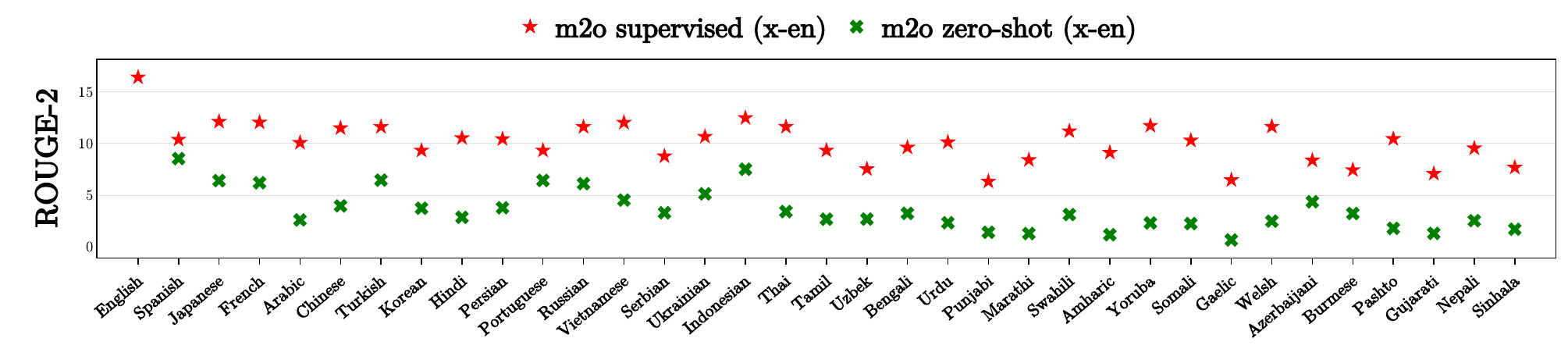}
\includegraphics[width=\textwidth]{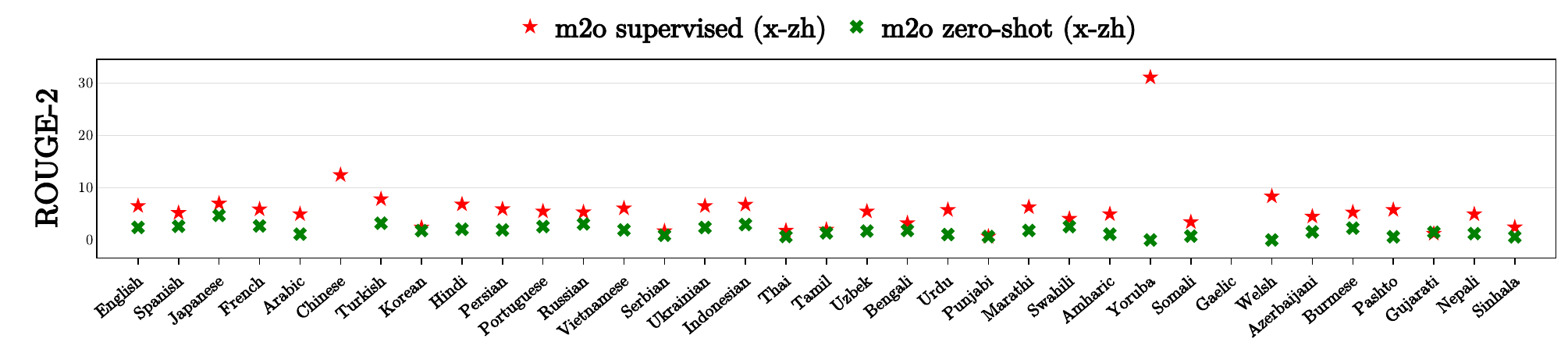}
\includegraphics[width=\textwidth]{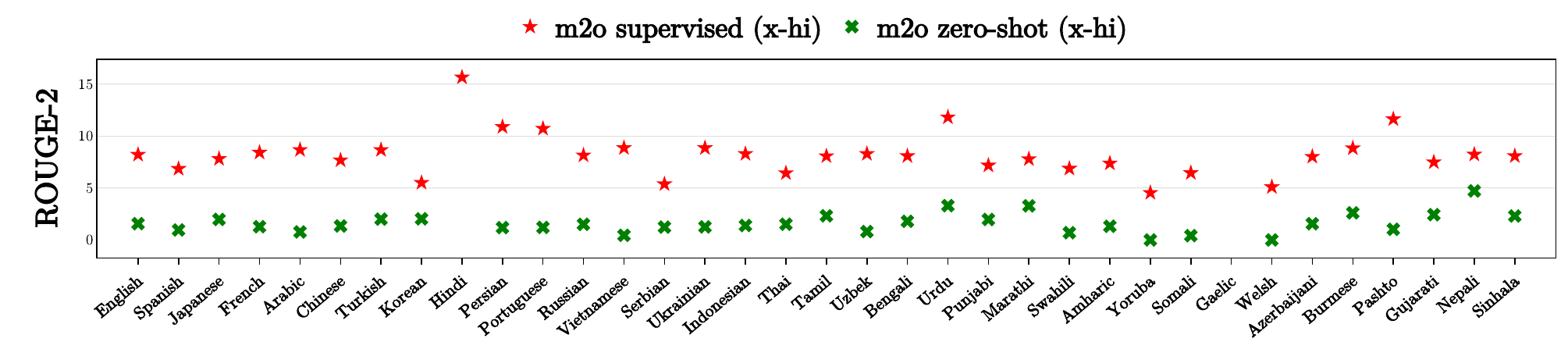}
\includegraphics[width=\textwidth]{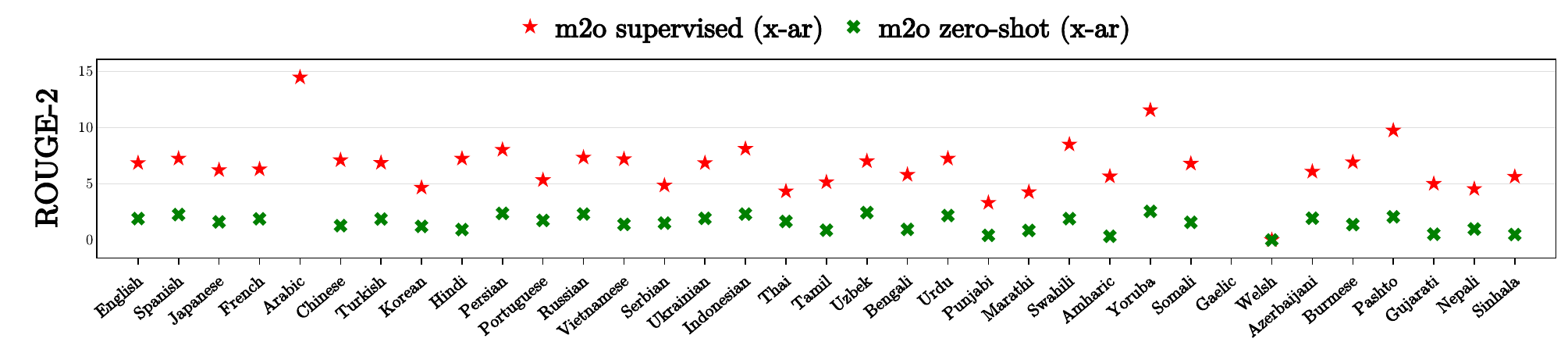}
\includegraphics[width=\textwidth]{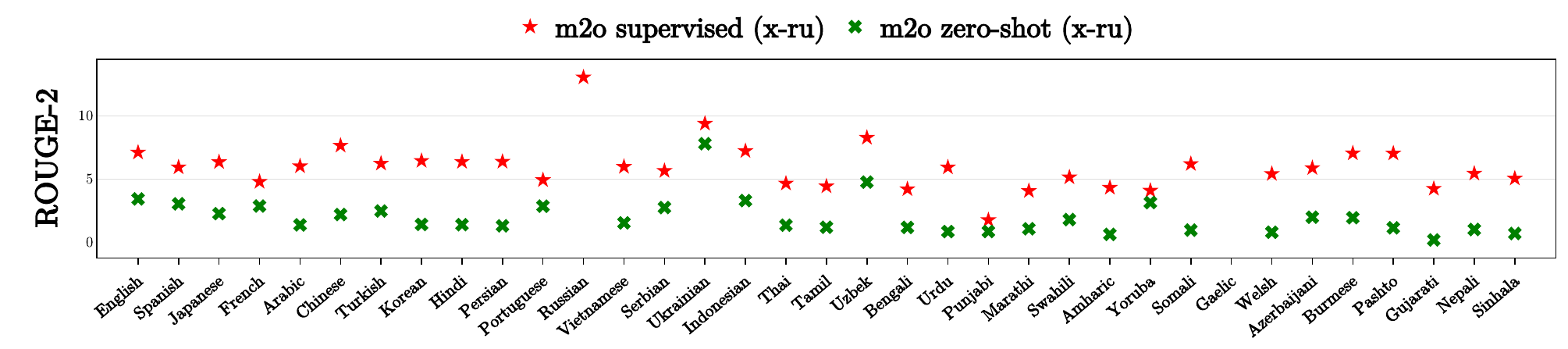}
\caption{Zero-shot ROUGE-2 scores for the different target languages as the source languages vary. The zero-shot models are trained with only the in-language samples of the pivot. Though their results are clearly behind the fully supervised models, the zero-shot models are able to generate non-trivial summaries for many language pairs.}
\label{fig:m2o-zs}
\end{figure*}

\begin{figure*}[t]
\centering
\includegraphics[width=\textwidth]{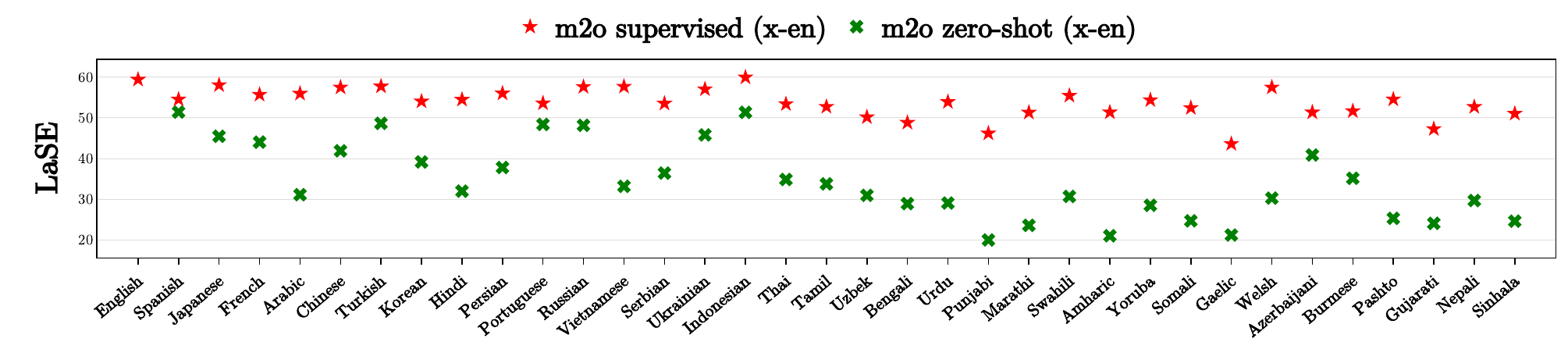}
\includegraphics[width=\textwidth]{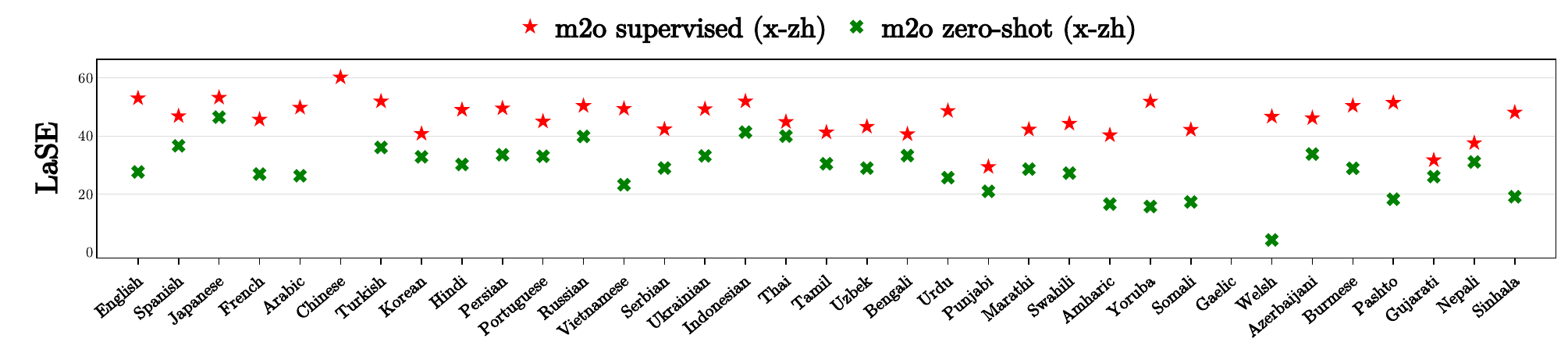}
\includegraphics[width=\textwidth]{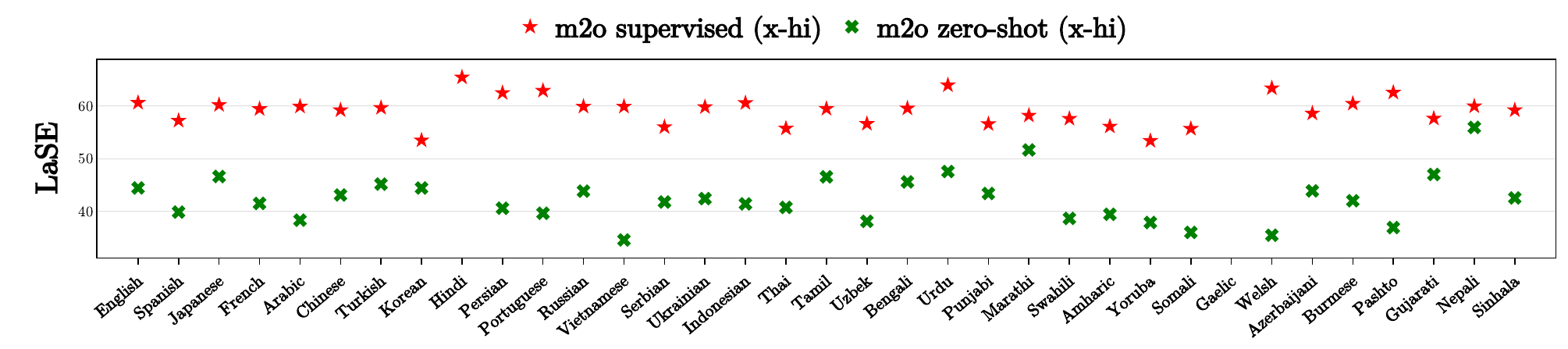}
\includegraphics[width=\textwidth]{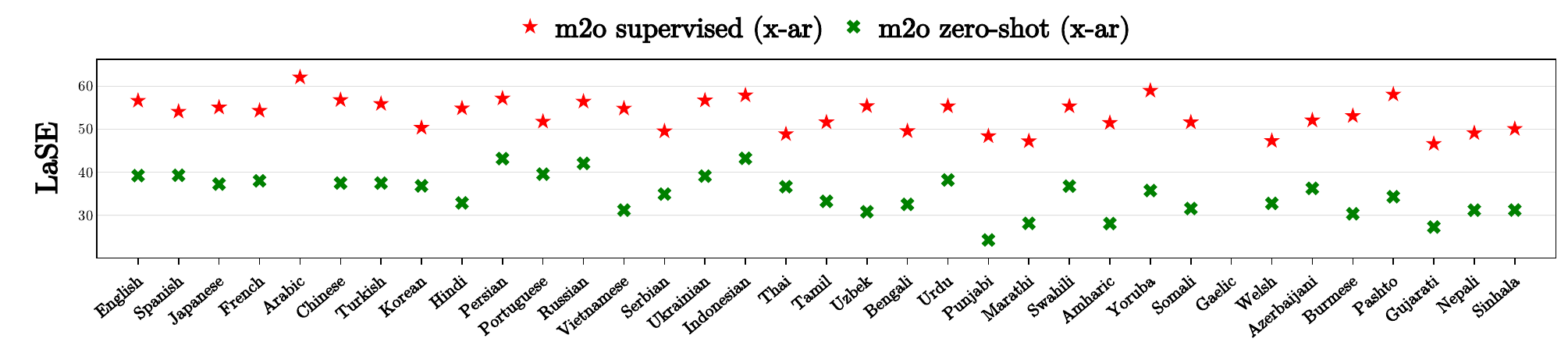}
\includegraphics[width=\textwidth]{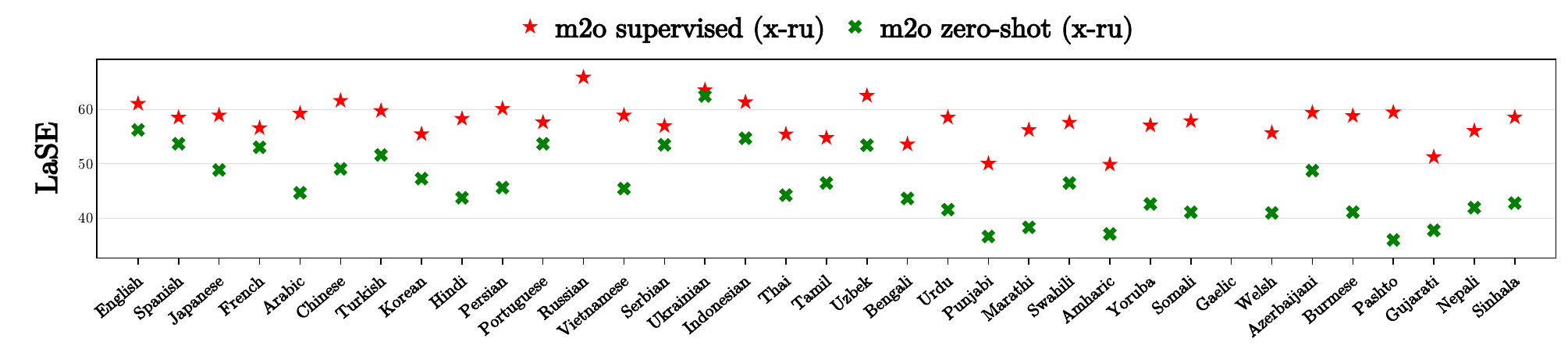}
\caption{Zero-shot LaSE scores for the different source languages as the target languages vary. The zero-shot models are trained with only the in-language samples of the pivot. Though their results are clearly behind the fully supervised models, the zero-shot models are able to generate non-trivial summaries for many language pairs.}
\label{fig:o2m-zs}
\end{figure*}

\begin{sidewaystable*}[t]
	\centering
	\setlength{\tabcolsep}{1.25pt}
    \tiny
\begin{tabular}[b]{c| ccccccccccccccccccccccccccccccccccccccccccccc |c}
				\toprule
				\bf{Language} & \bf{am} & \bf{ar} & \bf{az} & \bf{bn} & \bf{my} & \bf{zh-CN} & \bf{zh-TW} & \bf{en} & \bf{fr} & \bf{gu} & \bf{ha} & \bf{hi} & \bf{ig} & \bf{id} & \bf{ja} & \bf{rn} & \bf{ko} & \bf{ky} & \bf{mr} & \bf{ne} & \bf{om} & \bf{ps} & \bf{fa} & \bf{pcm} & \bf{pt} & \bf{pa} & \bf{ru} & \bf{gd} & \bf{sr-C} & \bf{sr-L} & \bf{si} & \bf{so} & \bf{es} & \bf{sw} & \bf{ta} & \bf{te} & \bf{th} & \bf{ti} & \bf{tr} & \bf{uk} & \bf{ur} & \bf{uz} & \bf{vi} & \bf{cy} & \bf{yo} & \bf{Total} \\

				\midrule
				\bf{am} & -- & 659 & 95 & 274 & 95 & 179 & 169 & 1445 & 371 & 171 & 220 & 361 & 31 & 497 & 269 & 415 & 239 & 93 & 223 & 304 & 19 & 189 & 423 & 205 & 291 & 191 & 333 & 0 & 350 & 361 & 62 & 299 & 346 & 383 & 374 & 322 & 122 & 129 & 424 & 341 & 393 & 40 & 287 & 1 & 71 & \bf{12066} \\
				\bf{ar} & 659 & -- & 781 & 799 & 646 & 2905 & 2783 & 9630 & 991 & 467 & 733 & 3651 & 83 & 6061 & 1175 & 873 & 691 & 302 & 547 & 844 & 9 & 2148 & 4170 & 427 & 2507 & 541 & 5329 & 1 & 1101 & 1139 & 316 & 1049 & 3650 & 1175 & 1294 & 852 & 371 & 29 & 4106 & 3429 & 4900 & 381 & 2623 & 39 & 141 & \bf{76348} \\
				\bf{az} & 95 & 781 & -- & 283 & 81 & 363 & 324 & 1307 & 203 & 181 & 124 & 735 & 26 & 1111 & 226 & 178 & 162 & 228 & 198 & 246 & 2 & 249 & 814 & 93 & 668 & 186 & 2087 & 3 & 286 & 285 & 124 & 359 & 704 & 535 & 505 & 233 & 139 & 2 & 1476 & 1373 & 957 & 195 & 726 & 31 & 40 & \bf{18924} \\
				\bf{bn} & 274 & 799 & 283 & -- & 145 & 308 & 275 & 1544 & 320 & 551 & 231 & 1376 & 37 & 1072 & 344 & 297 & 351 & 154 & 580 & 665 & 2 & 296 & 787 & 132 & 769 & 574 & 792 & 0 & 559 & 560 & 154 & 411 & 697 & 477 & 913 & 783 & 245 & 6 & 857 & 692 & 1381 & 96 & 521 & 35 & 62 & \bf{21407} \\
				\bf{my} & 95 & 646 & 81 & 145 & -- & 349 & 321 & 694 & 88 & 99 & 71 & 522 & 10 & 767 & 148 & 105 & 116 & 53 & 91 & 147 & 1 & 237 & 432 & 38 & 232 & 86 & 528 & 1 & 117 & 120 & 88 & 79 & 438 & 81 & 180 & 147 & 73 & 4 & 442 & 356 & 580 & 62 & 450 & 2 & 11 & \bf{9333} \\
				\bf{zh-CN} & 179 & 2905 & 363 & 308 & 349 & -- & 44561 & 4864 & 329 & 197 & 151 & 1331 & 34 & 2787 & 1010 & 227 & 407 & 135 & 236 & 269 & 13 & 552 & 1091 & 144 & 1334 & 235 & 2396 & 2 & 467 & 496 & 167 & 330 & 1941 & 402 & 500 & 352 & 263 & 13 & 1482 & 1591 & 1613 & 171 & 1853 & 28 & 40 & \bf{78118} \\
				\bf{zh-TW} & 169 & 2783 & 324 & 275 & 321 & 44561 & -- & 4777 & 307 & 167 & 135 & 1167 & 31 & 2573 & 955 & 208 & 384 & 125 & 205 & 248 & 15 & 499 & 947 & 134 & 1224 & 219 & 2166 & 1 & 418 & 457 & 160 & 302 & 1817 & 372 & 455 & 328 & 243 & 15 & 1273 & 1438 & 1420 & 162 & 1655 & 26 & 39 & \bf{75500} \\
				\bf{en} & 1445 & 9630 & 1307 & 1544 & 694 & 4864 & 4777 & -- & 1891 & 973 & 916 & 4668 & 147 & 10012 & 3035 & 1870 & 1686 & 497 & 1172 & 1600 & 35 & 1514 & 4717 & 1076 & 4714 & 1315 & 8680 & 127 & 3748 & 3798 & 525 & 2139 & 6891 & 2701 & 3134 & 2111 & 1014 & 58 & 5612 & 6530 & 6319 & 450 & 4580 & 2636 & 229 & \bf{127381} \\
				\bf{fr} & 371 & 991 & 203 & 320 & 88 & 329 & 307 & 1891 & -- & 227 & 476 & 607 & 105 & 1020 & 275 & 723 & 270 & 118 & 238 & 322 & 5 & 189 & 609 & 440 & 913 & 237 & 802 & 2 & 553 & 570 & 102 & 499 & 987 & 870 & 423 & 379 & 180 & 12 & 820 & 717 & 767 & 73 & 442 & 40 & 163 & \bf{19675} \\
				\bf{gu} & 171 & 467 & 181 & 551 & 99 & 197 & 167 & 973 & 227 & -- & 138 & 5087 & 37 & 706 & 217 & 180 & 263 & 101 & 2057 & 547 & 1 & 238 & 511 & 98 & 524 & 2161 & 550 & 1 & 337 & 339 & 132 & 256 & 532 & 307 & 1728 & 2020 & 162 & 5 & 616 & 506 & 1605 & 69 & 442 & 23 & 49 & \bf{25578} \\
				\bf{ha} & 220 & 733 & 124 & 231 & 71 & 151 & 135 & 916 & 476 & 138 & -- & 454 & 202 & 897 & 163 & 484 & 141 & 61 & 155 & 238 & 6 & 222 & 480 & 518 & 372 & 145 & 507 & 1 & 248 & 259 & 52 & 386 & 456 & 566 & 294 & 250 & 85 & 8 & 511 & 405 & 522 & 56 & 357 & 31 & 361 & \bf{13088} \\
				\bf{hi} & 361 & 3651 & 735 & 1376 & 522 & 1331 & 1167 & 4668 & 607 & 5087 & 454 & -- & 60 & 5598 & 619 & 479 & 509 & 231 & 3757 & 1340 & 3 & 1504 & 5293 & 187 & 6478 & 3971 & 4434 & 2 & 806 & 808 & 442 & 732 & 2917 & 896 & 3631 & 3696 & 367 & 9 & 3667 & 3912 & 15502 & 342 & 3706 & 80 & 77 & \bf{96014} \\
				\bf{ig} & 31 & 83 & 26 & 37 & 10 & 34 & 31 & 147 & 105 & 37 & 202 & 60 & -- & 116 & 23 & 105 & 28 & 17 & 52 & 40 & 5 & 9 & 48 & 251 & 62 & 39 & 79 & 0 & 45 & 48 & 12 & 72 & 87 & 151 & 56 & 50 & 16 & 5 & 92 & 74 & 60 & 11 & 61 & 6 & 291 & \bf{2814} \\
				\bf{id} & 497 & 6061 & 1111 & 1072 & 767 & 2787 & 2573 & 10012 & 1020 & 706 & 897 & 5598 & 116 & -- & 1271 & 986 & 784 & 348 & 755 & 1101 & 9 & 1450 & 3883 & 363 & 4375 & 718 & 7274 & 5 & 1377 & 1373 & 478 & 1303 & 4540 & 1873 & 1867 & 1129 & 603 & 11 & 5630 & 4799 & 6468 & 428 & 4790 & 146 & 172 & \bf{93526} \\
				\bf{ja} & 269 & 1175 & 226 & 344 & 148 & 1010 & 955 & 3035 & 275 & 217 & 163 & 619 & 23 & 1271 & -- & 368 & 660 & 143 & 298 & 417 & 3 & 270 & 1014 & 154 & 701 & 264 & 1419 & 2 & 555 & 568 & 112 & 388 & 950 & 426 & 631 & 420 & 307 & 4 & 1242 & 1016 & 806 & 54 & 901 & 22 & 31 & \bf{23876} \\
				\bf{rn} & 415 & 873 & 178 & 297 & 105 & 227 & 208 & 1870 & 723 & 180 & 484 & 479 & 105 & 986 & 368 & -- & 279 & 108 & 237 & 370 & 17 & 227 & 677 & 392 & 510 & 196 & 670 & 0 & 442 & 441 & 80 & 580 & 595 & 1183 & 507 & 351 & 146 & 13 & 709 & 609 & 614 & 55 & 613 & 19 & 173 & \bf{18311} \\
				\bf{ko} & 239 & 691 & 162 & 351 & 116 & 407 & 384 & 1686 & 270 & 263 & 141 & 509 & 28 & 784 & 660 & 279 & -- & 94 & 314 & 448 & 1 & 149 & 582 & 136 & 581 & 269 & 617 & 1 & 522 & 536 & 87 & 240 & 607 & 318 & 530 & 441 & 190 & 4 & 672 & 611 & 527 & 54 & 524 & 15 & 46 & \bf{16086} \\
				\bf{ky} & 93 & 302 & 228 & 154 & 53 & 135 & 125 & 497 & 118 & 101 & 61 & 231 & 17 & 348 & 143 & 108 & 94 & -- & 105 & 155 & 4 & 97 & 251 & 60 & 247 & 117 & 955 & 1 & 200 & 207 & 50 & 151 & 259 & 145 & 205 & 175 & 111 & 4 & 340 & 505 & 263 & 113 & 208 & 9 & 26 & \bf{7771} \\
				\bf{mr} & 223 & 547 & 198 & 580 & 91 & 236 & 205 & 1172 & 238 & 2057 & 155 & 3757 & 52 & 755 & 298 & 237 & 314 & 105 & -- & 617 & 2 & 228 & 604 & 137 & 532 & 1759 & 633 & 1 & 422 & 440 & 131 & 263 & 593 & 327 & 1746 & 1870 & 194 & 10 & 704 & 590 & 1381 & 75 & 473 & 15 & 50 & \bf{25017} \\
				\bf{ne} & 304 & 844 & 246 & 665 & 147 & 269 & 248 & 1600 & 322 & 547 & 238 & 1340 & 40 & 1101 & 417 & 370 & 448 & 155 & 617 & -- & 1 & 291 & 915 & 127 & 703 & 530 & 815 & 2 & 547 & 545 & 164 & 410 & 681 & 511 & 973 & 741 & 227 & 7 & 923 & 744 & 1154 & 81 & 714 & 31 & 66 & \bf{21821} \\
				\bf{om} & 19 & 9 & 2 & 2 & 1 & 13 & 15 & 35 & 5 & 1 & 6 & 3 & 5 & 9 & 3 & 17 & 1 & 4 & 2 & 1 & -- & 2 & 4 & 10 & 4 & 3 & 8 & 0 & 4 & 6 & 0 & 6 & 9 & 4 & 3 & 2 & 2 & 100 & 4 & 11 & 1 & 4 & 2 & 1 & 5 & \bf{348} \\
				\bf{ps} & 189 & 2148 & 249 & 296 & 237 & 552 & 499 & 1514 & 189 & 238 & 222 & 1504 & 9 & 1450 & 270 & 227 & 149 & 97 & 228 & 291 & 2 & -- & 2788 & 92 & 591 & 250 & 1213 & 0 & 220 & 231 & 146 & 305 & 763 & 314 & 435 & 308 & 90 & 7 & 1033 & 818 & 2812 & 160 & 657 & 7 & 33 & \bf{23833} \\
				\bf{fa} & 423 & 4170 & 814 & 787 & 432 & 1091 & 947 & 4717 & 609 & 511 & 480 & 5293 & 48 & 3883 & 1014 & 677 & 582 & 251 & 604 & 915 & 4 & 2788 & -- & 191 & 5461 & 523 & 4125 & 1 & 1011 & 1011 & 265 & 820 & 2532 & 1002 & 1223 & 775 & 363 & 8 & 3644 & 3542 & 6694 & 306 & 3167 & 68 & 73 & \bf{67845} \\
				\bf{pcm} & 205 & 427 & 93 & 132 & 38 & 144 & 134 & 1076 & 440 & 98 & 518 & 187 & 251 & 363 & 154 & 392 & 136 & 60 & 137 & 127 & 10 & 92 & 191 & -- & 229 & 106 & 306 & 0 & 240 & 247 & 30 & 220 & 315 & 428 & 219 & 154 & 88 & 26 & 279 & 284 & 227 & 19 & 174 & 7 & 462 & \bf{9465} \\
				\bf{pt} & 291 & 2507 & 668 & 769 & 232 & 1334 & 1224 & 4714 & 913 & 524 & 372 & 6478 & 62 & 4375 & 701 & 510 & 581 & 247 & 532 & 703 & 4 & 591 & 5461 & 229 & -- & 553 & 4247 & 7 & 1359 & 1343 & 232 & 612 & 7071 & 984 & 1034 & 806 & 472 & 4 & 3451 & 4374 & 6654 & 182 & 3732 & 110 & 96 & \bf{71345} \\
				\bf{pa} & 191 & 541 & 186 & 574 & 86 & 235 & 219 & 1315 & 237 & 2161 & 145 & 3971 & 39 & 718 & 264 & 196 & 269 & 117 & 1759 & 530 & 3 & 250 & 523 & 106 & 553 & -- & 589 & 2 & 399 & 399 & 126 & 288 & 566 & 356 & 1667 & 1854 & 195 & 11 & 615 & 562 & 1484 & 68 & 425 & 12 & 39 & \bf{24845} \\
				\bf{ru} & 333 & 5329 & 2087 & 792 & 528 & 2396 & 2166 & 8680 & 802 & 550 & 507 & 4434 & 79 & 7274 & 1419 & 670 & 617 & 955 & 633 & 815 & 8 & 1213 & 4125 & 306 & 4247 & 589 & -- & 4 & 1427 & 1413 & 354 & 1097 & 4652 & 1557 & 1526 & 849 & 557 & 9 & 5906 & 20706 & 5036 & 765 & 3759 & 131 & 115 & \bf{101417} \\
				\bf{gd} & 0 & 1 & 3 & 0 & 1 & 2 & 1 & 127 & 2 & 1 & 1 & 2 & 0 & 5 & 2 & 0 & 1 & 1 & 1 & 2 & 0 & 0 & 1 & 0 & 7 & 2 & 4 & -- & 2 & 3 & 2 & 1 & 3 & 1 & 1 & 1 & 1 & 0 & 6 & 4 & 3 & 0 & 4 & 36 & 2 & \bf{237} \\
				\bf{sr-C} & 350 & 1101 & 286 & 559 & 117 & 467 & 418 & 3748 & 553 & 337 & 248 & 806 & 45 & 1377 & 555 & 442 & 522 & 200 & 422 & 547 & 4 & 220 & 1011 & 240 & 1359 & 399 & 1427 & 2 & -- & 9000 & 124 & 375 & 1225 & 564 & 748 & 677 & 337 & 6 & 1248 & 1514 & 1013 & 109 & 674 & 43 & 72 & \bf{35491} \\
				\bf{sr-L} & 361 & 1139 & 285 & 560 & 120 & 496 & 457 & 3798 & 570 & 339 & 259 & 808 & 48 & 1373 & 568 & 441 & 536 & 207 & 440 & 545 & 6 & 231 & 1011 & 247 & 1343 & 399 & 1413 & 3 & 9000 & -- & 133 & 381 & 1258 & 560 & 768 & 688 & 345 & 9 & 1239 & 1506 & 1009 & 109 & 631 & 45 & 74 & \bf{35758} \\
				\bf{si} & 62 & 316 & 124 & 154 & 88 & 167 & 160 & 525 & 102 & 132 & 52 & 442 & 12 & 478 & 112 & 80 & 87 & 50 & 131 & 164 & 0 & 146 & 265 & 30 & 232 & 126 & 354 & 2 & 124 & 133 & -- & 132 & 259 & 186 & 345 & 172 & 71 & 6 & 302 & 309 & 512 & 39 & 217 & 8 & 14 & \bf{7422} \\
				\bf{so} & 299 & 1049 & 359 & 411 & 79 & 330 & 302 & 2139 & 499 & 256 & 386 & 732 & 72 & 1303 & 388 & 580 & 240 & 151 & 263 & 410 & 6 & 305 & 820 & 220 & 612 & 288 & 1097 & 1 & 375 & 381 & 132 & -- & 682 & 1024 & 712 & 373 & 172 & 17 & 955 & 874 & 1005 & 73 & 729 & 21 & 110 & \bf{21232} \\
				\bf{es} & 346 & 3650 & 704 & 697 & 438 & 1941 & 1817 & 6891 & 987 & 532 & 456 & 2917 & 87 & 4540 & 950 & 595 & 607 & 259 & 593 & 681 & 9 & 763 & 2532 & 315 & 7071 & 566 & 4652 & 3 & 1225 & 1258 & 259 & 682 & -- & 1045 & 1051 & 831 & 480 & 12 & 3617 & 3119 & 3046 & 287 & 2318 & 55 & 134 & \bf{65018} \\
				\bf{sw} & 383 & 1175 & 535 & 477 & 81 & 402 & 372 & 2701 & 870 & 307 & 566 & 896 & 151 & 1873 & 426 & 1183 & 318 & 145 & 327 & 511 & 4 & 314 & 1002 & 428 & 984 & 356 & 1557 & 1 & 564 & 560 & 186 & 1024 & 1045 & -- & 934 & 495 & 264 & 11 & 1350 & 1294 & 1243 & 81 & 928 & 35 & 216 & \bf{28575} \\
				\bf{ta} & 374 & 1294 & 505 & 913 & 180 & 500 & 455 & 3134 & 423 & 1728 & 294 & 3631 & 56 & 1867 & 631 & 507 & 530 & 205 & 1746 & 973 & 3 & 435 & 1223 & 219 & 1034 & 1667 & 1526 & 1 & 748 & 768 & 345 & 712 & 1051 & 934 & -- & 2236 & 388 & 12 & 1467 & 1414 & 2393 & 114 & 1069 & 32 & 72 & \bf{39809} \\
				\bf{te} & 322 & 852 & 233 & 783 & 147 & 352 & 328 & 2111 & 379 & 2020 & 250 & 3696 & 50 & 1129 & 420 & 351 & 441 & 175 & 1870 & 741 & 2 & 308 & 775 & 154 & 806 & 1854 & 849 & 1 & 677 & 688 & 172 & 373 & 831 & 495 & 2236 & -- & 306 & 11 & 875 & 832 & 1743 & 99 & 634 & 20 & 62 & \bf{31453} \\
				\bf{th} & 122 & 371 & 139 & 245 & 73 & 263 & 243 & 1014 & 180 & 162 & 85 & 367 & 16 & 603 & 307 & 146 & 190 & 111 & 194 & 227 & 2 & 90 & 363 & 88 & 472 & 195 & 557 & 1 & 337 & 345 & 71 & 172 & 480 & 264 & 388 & 306 & -- & 3 & 469 & 482 & 424 & 33 & 355 & 13 & 23 & \bf{10991} \\
				\bf{ti} & 129 & 29 & 2 & 6 & 4 & 13 & 15 & 58 & 12 & 5 & 8 & 9 & 5 & 11 & 4 & 13 & 4 & 4 & 10 & 7 & 100 & 7 & 8 & 26 & 4 & 11 & 9 & 0 & 6 & 9 & 6 & 17 & 12 & 11 & 12 & 11 & 3 & -- & 9 & 9 & 5 & 2 & 4 & 0 & 6 & \bf{635} \\
				\bf{tr} & 424 & 4106 & 1476 & 857 & 442 & 1482 & 1273 & 5612 & 820 & 616 & 511 & 3667 & 92 & 5630 & 1242 & 709 & 672 & 340 & 704 & 923 & 4 & 1033 & 3644 & 279 & 3451 & 615 & 5906 & 6 & 1248 & 1239 & 302 & 955 & 3617 & 1350 & 1467 & 875 & 469 & 9 & -- & 4085 & 4314 & 361 & 2953 & 127 & 128 & \bf{70035} \\
				\bf{uk} & 341 & 3429 & 1373 & 692 & 356 & 1591 & 1438 & 6530 & 717 & 506 & 405 & 3912 & 74 & 4799 & 1016 & 609 & 611 & 505 & 590 & 744 & 11 & 818 & 3542 & 284 & 4374 & 562 & 20706 & 4 & 1514 & 1506 & 309 & 874 & 3119 & 1294 & 1414 & 832 & 482 & 9 & 4085 & -- & 4252 & 438 & 2992 & 105 & 92 & \bf{83856} \\
				\bf{ur} & 393 & 4900 & 957 & 1381 & 580 & 1613 & 1420 & 6319 & 767 & 1605 & 522 & 15502 & 60 & 6468 & 806 & 614 & 527 & 263 & 1381 & 1154 & 1 & 2812 & 6694 & 227 & 6654 & 1484 & 5036 & 3 & 1013 & 1009 & 512 & 1005 & 3046 & 1243 & 2393 & 1743 & 424 & 5 & 4314 & 4252 & -- & 391 & 3707 & 70 & 85 & \bf{95355} \\
				\bf{uz} & 40 & 381 & 195 & 96 & 62 & 171 & 162 & 450 & 73 & 69 & 56 & 342 & 11 & 428 & 54 & 55 & 54 & 113 & 75 & 81 & 4 & 160 & 306 & 19 & 182 & 68 & 765 & 0 & 109 & 109 & 39 & 73 & 287 & 81 & 114 & 99 & 33 & 2 & 361 & 438 & 391 & -- & 259 & 11 & 18 & \bf{6896} \\
				\bf{vi} & 287 & 2623 & 726 & 521 & 450 & 1853 & 1655 & 4580 & 442 & 442 & 357 & 3706 & 61 & 4790 & 901 & 613 & 524 & 208 & 473 & 714 & 2 & 657 & 3167 & 174 & 3732 & 425 & 3759 & 4 & 674 & 631 & 217 & 729 & 2318 & 928 & 1069 & 634 & 355 & 4 & 2953 & 2992 & 3707 & 259 & -- & 101 & 78 & \bf{55495} \\
				\bf{cy} & 1 & 39 & 31 & 35 & 2 & 28 & 26 & 2636 & 40 & 23 & 31 & 80 & 6 & 146 & 22 & 19 & 15 & 9 & 15 & 31 & 1 & 7 & 68 & 7 & 110 & 12 & 131 & 36 & 43 & 45 & 8 & 21 & 55 & 35 & 32 & 20 & 13 & 0 & 127 & 105 & 70 & 11 & 101 & -- & 8 & \bf{4301} \\
				\bf{yo} & 71 & 141 & 40 & 62 & 11 & 40 & 39 & 229 & 163 & 49 & 361 & 77 & 291 & 172 & 31 & 173 & 46 & 26 & 50 & 66 & 5 & 33 & 73 & 462 & 96 & 39 & 115 & 2 & 72 & 74 & 14 & 110 & 134 & 216 & 72 & 62 & 23 & 6 & 128 & 92 & 85 & 18 & 78 & 8 & -- & \bf{4155} \\

				\midrule
				\bottomrule
			\end{tabular}
   
	\caption{
	An article-summary statistics of the CrossSum dataset containing a total of 1,678,466 cross-lingual samples. The rows indicate the articles' language, and the columns of their summaries'. For example, the cell on the second column of the fourth row indicates the number of samples where the article is in Bengali and the summary in Arabic.}
	\label{tab:crosssum}
	\centering
\end{sidewaystable*}

\section{Modeling Details}\label{sec:modeling_details}

\subsection{Choice of Pretrained Model}

Many pretrained multilingual text-to-text models are currently available, e.g., mBART \citep{liu2020multilingual}, CRISS \citep{NEURIPS2020_1763ea5a}, MARGE \citep{lewis2020marge}, and mT5 \citep{xue2020mt5}. While mBART and mT5 are pretrained with multilingual objectives, CRISS and MARGE are pretrained with a cross-lingual one, which better suits our use case. However, we choose mT5 for fine-tuning because of its broad coverage of 101 languages with support for 41 of the 45 languages from CrossSum, in contrast to only 15 languages in mBART or CRISS and 26 in MARGE.

\subsection{Summarize-then-translate (s. + t.)} The primary reason for using summarize-then-translate rather than translate-then-summarize is the computational cost between these two. Available translation models only work for short sequences and are unsuitable for long documents. One solution is to segment the documents into sentences and then translate them. But that increases the compute overhead, and translations suffer from loss of context. We use a multilingual summarization model \citep{hasan-etal-2021-xl} coupled with the multilingual machine translation model, M2M-100 \citep{fan2021beyond}, for our pipeline. 

\subsubsection{Multilingual Summarization} The pipeline first performs in-language summarization. We train our own model for summarization as the model released by \citet{hasan-etal-2021-xl} has been rendered unusable due to the change in the dataset split. We extend our component graphs to curate the in-language dataset splits. We consider articles having no parallel counterpart in any other language as single node components in the component graph. As before, we assign all articles originating from a single component to the training (dev/test) set of the dataset, extending them to the in-language splits too. We then train the multilingual model by fine-tuning mT5 with the in-language splits, sampling each batch of 256 samples from a single language with a sampling factor of $\alpha=0.5$.

\subsubsection{Multilingual Translation} For multilingual translation, we used M2M-100 \citep{fan2021beyond} (418M parameters variant), a many-to-many multilingual translation model, with support for 37 languages from CrossSum.

\subsection{Many-to-One (m2o) Model} Many-to-one training is standard for evaluating cross-lingual summarization. In these models, the language of the source text can vary, but the target language remains the same, i.e., as the pivot language. Instead of sampling all samples of a batch from the same language pair, we sample 8 mini-batches of 32 samples using a sampling factor of $\alpha = 0.25$, the source side of each originating from a single language while the target language remains fixed. We then merge the mini-batches into a single batch and update the model parameters. This is to ensure that there are not many duplicates in a single batch (if all 256 samples of a batch are sampled from a single language pair, there might be many duplicates as many language pairs do not have 256 training samples) and the model still benefits the advantages of low-resource upsampling.

\subsection{One-to-many (o2m) Model} o2m models are complementary to m2o models: we train them by keeping the source language fixed and varying the target language. We upsample the low-resource target languages with the same sampling factor of $\alpha = 0.25$ and merge 8 mini-batches of 32 samples each, analogous to m2o models.

\subsection{Many-to-many (m2m) Multistage Model}

This is the model obtained from the Algorithm \ref{alg:samp}. In contrast to standard language sampling \citep{conneau2019unsupervised}, we sample the target language and then choose the source based on that decision. We use batch size 256, 8 mini-batches with size 32, and $\alpha = 0.5, \beta = 0.75$. 

\subsection{Many-to-many (m2m) Unistage Model}

This algorithm is similar to standard language sampling, the difference being that languages are sampled as pairs from all possible combinations. Instead of sampling one language pair at each training step, we sample 8 pairs, one for each mini-batch of size 32. We then merge the mini-batches into a single batch of 256 samples before updating the model parameters. We use a sampling factor of $\alpha = 0.25$.

In all models, we discarded a language pair from training if it had fewer than 30 training samples to prevent too many duplicates in a mini-batch. The training was done together with the in-language samples.

\section {Experimental Details}

\subsection{Training Setups}\label{sec:training_setups}

Fine-tuning generation models is compute-intensive, and due to computational limitations, we fine-tune all pretrained models for 25k steps with an effective batch size of 256, which roughly takes about three days on a 4-GPU NVIDIA P100 server. We use the base variant of mT5, having 250k vocabulary, 768 embedding and dimension size, 12 attention heads, and 2048 FFN size,  with 580M parameters. We limit the input to 512 and output to 84 tokens. All models are trained on the respective subsets of the CrossSum training set.

\subsection{Inference} During inference, we jump-start the decoder with language-specific \texttt{BOS} (beginning of sequence) tokens \citep{johnson2017google} at the first decoding step for guiding the decoder to generate summaries in the intended target language. We use beam search \citep{medress1977speech} with the beam size 4 and use a length penalty \citep{wu2016google} of 0.6. 

\section{Ablation Studies}\label{sec:ablation}

We make several design choices in the multistage sampling algorithm. We break them into two main decisions:\begin{enumerate}
    \item Making mini-batches and sampling the language pair for each mini-batch.
    \item Keeping either the source or the target language fixed for each batch.  
\end{enumerate}

To verify that these choices indeed affect performance positively, we train five different models for ablation:\begin{enumerate}
    \item Sampling the language pair in mini-batches in one stage only and then merging them into large batches before updating model parameters: m2m-unistage.
    \item Sampling the language pair with large batches of 256 samples without mini-batching: m2m-large.
    \item Multistage sampling keeping only the target language fixed in a batch: m2m-tgt \textit{[our proposed model]}.
    \item Multistage sampling keeping only the source language fixed in a batch: m2m-src; i.e., the complement of our proposed model.
    \item Multistage sampling keeping either the source or the target language fixed (with equal probability) for each batch: m2m-src-tgt.
\end{enumerate}

We benchmark on all the language pairs done previously and show the mean ROUGE-2 and LaSE scores in Table \ref{tab:ablation}.

\begin{table}[H]
\centering\setlength{\tabcolsep}{2pt}
{%
\begin{tabular}{lcccc}
\hline
\multirow{2}{*}{Model} & Scores & \multicolumn{3}{c}{Significance}\\
\cline{2-5}
 & \footnotesize{R-2/LaSE} & \footnotesize{Better} & \footnotesize{Worse} & \footnotesize{Insignificant}\\
\hline
m2m-large & \textbf{8.31}/\textbf{57.45} & 122 & \textbf{59} & 503 \\
m2m-unistage & 7.51/55.36 & 191 & 149 & 344 \\
m2m-tgt & 8.15/57.15 & \textbf{289} & 66 & \textbf{329} \\
m2m-src & 4.44/26.75 & 34 & 477 & 173 \\
m2m-src-tgt & 6.47/42.55 & 89 & 297 & 298 \\
\hline
\end{tabular}
}
\caption{ROUGE-2 and LaSE scores for ablation.}
\label{tab:ablation}
\end{table}

As can be seen from the table, m2m-large, the standard m2m model, has the best average ROUGE-2/LaSE scores among all m2m variants. This begs the question of whether our proposed multistage sampling is, after all, needed or not. But the scores of the proposed m2m-tgt model do not fall much below. Therefore, we show statistical significance test results of all m2m models, comparing them against m2o, o2m, and s.+t. in one vs. all manner. 

Significance results paint a different picture: m2m-tgt triumphs over all other models, getting significantly better results on 42\% language pairs, more than double the m2m-large model. We inspected the results individually and found that the results are notably better on language pairs that are not adequately represented in the training set. m2m-tgt performs comparatively worse on high-resource language pairs, which we think is a fair compromise to uplift low-resource ones. As m2m-large can sample a pair only once per batch, it fails to incorporate many language pairs due to them having insufficient participation during training. On the other hand, our proposed multistage sampling algorithm performs well in this regard by sampling in two stages.

While m2m-tgt outperforms all the rest, m2m-src falls behind all other models by a large margin. This phenomenon also has the same trend as the results in Section \ref{sec:exp}, where o2m models failed at generating cross-lingual summaries. This is also in line with our hypothesis made, as m2m-src and m2m-tgt mimic the training settings of the o2m and m2o models, respectively, at the batch level. The m2m-src-tgt is the middle ground between m2m-src and m2m-tgt and, likewise, scores between these two. In our opinion, the performance dynamics between the m2o (m2m-tgt) and o2m (m2m-src) models is an interesting finding and should be studied in depth as a new research direction in future works.

\appendix



\end{document}